%% file: iclr2027_conference.tex
\documentclass{article} %
\usepackage{iclr2027_conference,times}

\input{math_commands.tex}

\usepackage{hyperref}
\usepackage{url}
\usepackage{graphicx}
\usepackage{amsmath,amssymb}
\usepackage{algorithm}
\usepackage{algpseudocode}
\usepackage{booktabs}
\usepackage{multirow}
\usepackage{makecell}
\usepackage{caption}
\usepackage{subcaption}
\usepackage{booktabs}
\usepackage{mathtools}
\usepackage{wrapfig}
\usepackage{enumitem}

\title{COMPASS: Membership Inference Attack Against Vision Large Language Models via Adaptive Noise Calibration}

\author{Damien Lo\thanks{Equal contribution.} ,
Hong Kyu Lee\footnotemark[1] ,
Ruixuan Liu ,
Li Xiong\\
Department of Computer Science\\
Emory University\\
Atlanta, GA\\
\texttt{\{damien.lo,hong.kyu,ruixuan.liu,lxiong\}@emory.edu}
}

\iclrfinalcopy %
\begin{document}

\maketitle

\input{sec/0_abstract}

\input{sec/1_intro}

\input{sec/2_related_work}

\input{sec/3_problem_setting}
\input{sec/4_the_low_std_region}

\input{sec/5_attack_methodology}

\input{sec/6_experiments}

\input{sec/7_conclusion}

\section*{AI use statement}
We used generative AI for searching related works, ideation and proving mathematical claims. Authors have rigorously analyzed the related manuscripts retrieved by AI. The research idea, objective and proposal are original to authors and was not assisted by AI. AI was used to validate details of the proposal and responses of AI was reveiwed and evaluated by authors. Mathematical analysis is original to authors, and AI was used to proofread our analysis. Responses from the mode has been rigorously analyzed.

\section*{Ethics Statement}
\label{subsec: Ethics Statement}
This work studies membership inference attacks to assess privacy risks in publicly available vision--language models, including a medical model. Such attacks could also be misused to infer whether an individual's data was included in training. We conduct our evaluations using publicly available models and datasets, and present COMPASS as a tool for privacy auditing and for informing the development of stronger protections. We have read and comply with the ICLR Code of Ethics.

\section*{Reproducibility Statement}
\label{subsec: Reproducibility Statement}
We use publicly available models and datasets, with links provided in \autoref{app:Assets} and \autoref{app:Experimental Setup}. We additionally provide our source code through the anonymized repository linked in \autoref{app:Assets}. The paper describes the COMPASS scoring and calibration methods, while \autoref{app:algorithms} gives their full algorithms. We also document the endoscopy target-set filtering procedure in \autoref{app:blind_classifier_filtering} and report the hardware and runtime requirements in \autoref{app:Experimental Setup}.

\bibliography{iclr2027_conference}
\bibliographystyle{iclr2027_conference}

\clearpage
\appendix
\raggedbottom

\input{sec/8_appendix}

\end{document}

%% file: math_commands.tex
\usepackage{amsmath,amsfonts,bm}

\newcommand{\partitle}[1]{\noindent\textbf{#1}}

\def\eqref#1{equation~\ref{#1}}

\def\1{\bm{1}}

\DeclareMathAlphabet{\mathsfit}{\encodingdefault}{\sfdefault}{m}{sl}
\SetMathAlphabet{\mathsfit}{bold}{\encodingdefault}{\sfdefault}{bx}{n}

%% file: sec/0_abstract.tex
\begin{abstract}
Vision large language models (VLLMs) are increasingly deployed in privacy-sensitive domains, raising concerns about the exposure of their training data. The strongest existing gray-box membership inference attacks (MIAs) against VLLMs commonly rely on entropy statistics or generated descriptions, requiring costly response generation while providing limited membership separation. In contrast, perturbation-based methods measure how model response changes when images are perturbed, but have mainly focused on large perturbations, leaving VLLM membership behavior under smaller perturbations poorly understood. In this work, we systematically characterize VLLM membership behavior under controlled Gaussian perturbations across the full range of noise levels at fine resolution. Our analysis reveals that both the optimal perturbation magnitude and the direction of the membership signal vary across model--dataset pairs. For many settings, a previously unstudied low-noise regime produces stronger separation than large perturbations, with members exhibiting lower divergence than non-members and thereby reversing the conventionally observed score direction. Other settings retain the conventional direction, demonstrating that neither a fixed perturbation magnitude nor a predetermined score orientation is universally reliable. 
Motivated by this finding, we propose \textbf{COMPASS}, an adaptive gray-box MIA that calibrates both the perturbation magnitude and membership-score direction using a loosely distribution-matched, known non-member reference set. We instantiate COMPASS using R\'enyi-normalized KL divergence and R\'enyi divergence, with max-$k$ aggregation over the most informative image-logit positions. Across LLaVA v1.5--7B, MiniGPT-4, and HuluMed-32B, spanning pretraining, instruction-tuning, and medical-image datasets, COMPASS consistently outperforms baselines. Under general-domain evaluations, COMPASS achieves AUCs of up to $0.88$. On the distribution-matched medical evaluation, it reaches $0.64$ AUC, while the strongest baseline remains at chance-level. These results establish full-range adaptive perturbation analysis as an effective approach to VLLM membership inference.
\end{abstract}

%% file: sec/1_intro.tex
\section{Introduction}
\label{sec:Introduction}

Vision large language models (VLLMs) have demonstrated exceptional ability on multi-modal tasks such as 
image question answering~\cite{hartsock2024vision, khan2024consistency}, medical 
diagnosis~\cite{bazi2023vision, guo2024prompting}, and reasoning problems ~\cite{nagar2024zero, chen2024spatialvlm, dong2025insight}. This rapid advancement raises concerns about data privacy, especially when these models are trained on private or copyrighted images~\cite{samson2024little}. VLLMs can memorize these data within their parameters and expose them through inference-time behavior~\cite{jayaraman2024deja}. The image modality is particularly concerning, as training images can be easily reconstructed from VLLMs~\cite{nguyen2025model}, and the image itself is sufficient to extract private attributes from the model~\cite{tomekcce2024private}. 

The membership inference attack (MIA)~\cite{carlini2022membership, fu2024membership,zhang2024min} serves as a practical way to verify whether private or copyrighted data was used during training~\cite{duarte2024cop}, and is widely used for privacy auditing~\cite{liu2024please} and verifying machine unlearning~\cite{shi2024muse}. Given a dataset containing both members (training samples) and non-members, MIA aims to distinguish members from non-members.
Existing attacks on VLLMs have important limitations: Hu~\textit{et al.} proposed an attack using a statistical test on generated responses~\cite{hu2025membership}, but it requires both modalities and determines only group-level membership for the vision modality. Li~\textit{et al.} use the entropy of responses as a membership signal~\cite{li2024membership}, at the cost of efficiency, since it requires full response generation.

Perturbation-based attacks alternatively apply a change to the input, measuring how much the model output shifts. The magnitude of this shift indicates how stably the model behaves around that sample. This sensitivity constitutes a natural membership signal, as training and unseen samples respond differently to the same change. Prior works have demonstrated the effectiveness of perturbation on vision classifiers~\cite{jayaraman2021revisiting,choquettechoo2021labelonly,li2021membership} and language models~\cite{mattern2023membership}, but has primarily considered strong image transformations, leaving behavior under small, controlled perturbations unexplored. Although \citet{li2024membership} evaluate the perturbation-based EncoderMI baseline~\cite{liu2021encodermi} on VLLMs, its augmentations serve as fixed preprocessing operations rather than an object of analysis: they are nonlinear and semantically nontrivial, their strength is not varied,
and the work provides no analysis of how the model's behavior differs between members and non-members. 
\cite{he2025towards} underscore the cost of this omission, showing that overly coarse perturbation causes robustness-based attacks on LLMs to fail. 

We therefore systematically characterize VLLM membership behavior under controlled Gaussian perturbations across the full range of noise levels at fine resolution. This analysis reveals that both the perturbation magnitude yielding the strongest membership separation and the direction of the membership signal vary across model–dataset pairs. In many settings, a previously unexplored low-noise regime produces substantially stronger separation than large perturbations, with members exhibiting lower divergence than non-members; in others, the conventional direction persists. These findings show that neither perturbation magnitude nor score orientation can be fixed a priori.

Motivated by this finding, we propose \textbf{COMPASS} (\textbf{C}alibrated \textbf{O}rientation and \textbf{M}agnitude \textbf{P}erturbation \textbf{A}ttack via \textbf{S}igma \textbf{S}weeps), an adaptive gray-box MIA that searches the full Gaussian-noise range. COMPASS compares the aggregate divergence response of the target set against a loosely distribution-matched non-member reference set, selects the noise level producing the strongest observable separation, and determines whether lower or higher divergence indicates membership. This procedure requires no labeled members and only accesses image logits from the original and perturbed inputs, avoiding response generation. We instantiate COMPASS as COMPASS-KL, using R\'enyi normalization followed by KL divergence, and COMPASS-R, using R\'enyi divergence directly. Both variants apply max-$k$ aggregation over the most responsive image-logit positions.

We evaluate COMPASS on LLaVA v1.5--7B, MiniGPT-4, and HuluMed-32B across pretraining, instruction-tuning, and medical-image datasets. COMPASS consistently outperforms existing entropy- and augmentation-based MIA baselines, while variation across target settings further validates the need for adaptive calibration.

\partitle{Contributions.}
Our contributions are summarized as follows:
\begin{enumerate}[itemsep=10pt, parsep=0pt, topsep=4pt, partopsep=0pt]
\item We conduct the first systematic full-range analysis of Gaussian perturbation behavior for VLLM membership inference, identifying %
a previously overlooked low-noise regime with an inverted membership signal  that provides the strongest separation in most evaluated settings. More broadly, we find that both the optimal perturbation magnitude and membership-signal direction vary across model–dataset pairs.
\item We propose COMPASS-KL and COMPASS-R, two adaptive gray-box MIA variants that calibrate perturbation magnitude and score direction using the target set and a loosely distribution-matched non-member reference set, without requiring labeled members. We instantiate COMPASS with two divergence measures, yielding COMPASS-KL and COMPASS-R.
\item Through extensive experiments across multiple VLLMs and training-data settings, we demonstrate that COMPASS consistently outperforms existing MIA baselines.
\end{enumerate}

%% file: sec/2_related_work.tex
\section{Related Work}
\label{sec:2_related_work}

\partitle{Vision Large Language Models (VLLMs)} 
Multi-modal models offer strong cross-modal understanding and are widely used in various domains~\cite{kumar2025medvisionllama,maiti2025gen3deval}.
Unlike earlier vision language models that achieve cross-modal understanding via contrastive alignment~\cite{qi2020imagebert,radford2021learning,jia2021scaling}, VLLMs project image representations into an LLM. The projector can be a transformer~\cite{li2022blip}, a linear layer~\cite{liu2023visual}, a multi-layer perceptron~\cite{zhu2023minigpt}, or cross-attention layers~\cite{bai2023versatile}.
COMPASS specifically targets VLLMs and assumes an LLM backbone in the target model. VLLMs have two training stages: (1) a pre-training stage that aligns the visual encoder and LLM backbone on paired image–text data; and (2) an instruction-tuning stage that optimizes the model to follow multi-modal instructions. Our attack aims to infer membership for datasets from both stages using the instruction-tuned model.

\partitle{Membership Inference Attack (MIA)}
Given a sample and a target model, MIA aims to determine whether the sample was part of the model's training set. Earlier work by Shokri~\textit{et al.} trains shadow models as a proxy for the target model and trains a binary classifier over their outputs~\cite{shokri2017membership}. Yeom \textit{et al.} proposed the loss-based attack~\cite{yeom2018privacy}, which determines a sample's membership from its loss under the model. Carlini~\textit{et al.} proposed LiRA, an MIA based on a likelihood ratio over a fitted distribution~\cite{carlini2022membership}. LiRA scales poorly to large models due to the prohibitive cost of training shadow models.
Reference-based calibration attacks~\cite{mireshghallah2022quantifying, carlini2021extracting, fu2024membership} offer an efficient proxy by capturing the loss of neighboring samples or of a reference model.
In parallel, Min-$k$\% detects training data using the average log-likelihood of the $k$\% of tokens with the lowest probabilities~\cite{shi2023detecting}. Normalizing the per-token prediction probabilities by the average probabilities amplifies the membership signal~\cite{zhang2024min}.

MIA has been studied for multi-modal models~\cite{ko2023practical, hu2022m,hu2025membership,li2024membership}. M$^4$I~\cite{hu2022m} targets VLMs with contrastive alignment. The attack model is a binary classifier trained on cross-modality features of members and non-members obtained from shadow models. This is not applicable to VLLMs, which have an LLM backbone. Hu~\textit{et al.} proposed the first membership inference attack on VLLMs. For each input sample, it generates a sequence from the model and conducts a likelihood ratio test~\cite{hu2025membership}. While effective, the attack requires a complete image--text pair. Although the paper also proposes an image-only attack, it can infer membership only for a group of images, not for individual images. Li~\textit{et al.} proposed a metric-based attack that leverages the average R\'enyi entropy over the output logits, where members show lower entropy than non-members~\cite{li2024membership}. However, the scope of the attack is limited to detecting pre-training data in the instruction-tuned model.

Adversarial perturbations~\cite{moosavi2017universal} have also been used for MIAs. Earlier works target classification models by directly launching an adversarial attack on the target samples~\cite{del2022leveraging, zhang2022evaluating}. The amount of noise serves as a membership signal, with members requiring larger perturbations. However, these attacks require white-box access. COMPASS is more analogous to label-only MIAs~\cite{choquettechoo2021labelonly,he2025towards}, where the adversary operates in a black-box setting and observes how the output changes under input perturbations. Recently, Liu~\textit{et al.} proposed a label-only MIA for VLMs~\cite{yihaolomia}. However, that attack requires a reference VLM and targets both modalities, so we do not compare against it.

%% file: sec/3_problem_setting.tex
\section{Problem Setting}
\label{sec:3_problem_setting}

Let $\mathcal{M}$ be a target VLLM trained on a dataset $X_{\mathrm{tr}}$. $\mathcal{M}$ takes an image--text pair $(x, z)$ as input, where $z$ is optional. $\mathcal{M}$ produces output logits $y\coloneq \left(y_1, \cdots, y_t\right)$, where $t$ is the sequence length. A target dataset $X \coloneqq X_m \cup X_n$ where $X_m \coloneqq\{x_m \vert x_m \in X_{\mathrm{tr}}\}$ and $X_n \coloneqq \{x_n\vert x_n \notin X_{\mathrm{tr}}\}$ denote the member and non-member sets, respectively.

Given a target set with unknown membership, the adversary's goal is to classify whether a given sample $x \in X$ belongs to $X_m$ or $X_n$. Formally, the objective is described as a binary classification problem with a threshold $\lambda$. Let $R(x,\mathcal{M})$ denote a raw attack statistic and let $o\in{-1,+1}$ denote its selected score orientation. We define the oriented membership score as
\begin{align}
S(x,\mathcal{M}) = o \cdot R(x,\mathcal{M}),
\end{align}
where $o$ is chosen such that larger values of $S$ indicate stronger evidence of membership. The resulting binary decision rule is
\begin{align*}
    \mathcal{A}\left(\mathbf{x}, \mathcal{M}\right) = \begin{cases}
        1 \quad\left(\mathbf{x} \in X_m\right)  & \text{if } S\left(\mathbf{x}, \mathcal{M}\right) \geq \lambda  \\
        0 \quad\left(\mathbf{x} \in X_n\right) & \text{otherwise}
    \end{cases}
\end{align*}
where a positive prediction flags $\mathbf{x}$ as a member. We assume a black-box setting, in which the attacker has access only to the input and the output logits of the model. We assume that the adversary has access to a public dataset that does not contain any members and follows a distribution similar to that of the target set. This is a realistic assumption as the attacker can use a publicly available dataset that is released after the model's training cutoff~\cite{hu2025membership}.

%% file: sec/4_the_low_std_region.tex
\section{Gaussian Perturbation Analysis}
\label{sec:4_the_low_std_region}
To characterize how VLLM membership behavior varies with perturbation magnitude, we systematically sweep additive Gaussian noise over a finely resolved range of standard deviations. This analysis reveals three distinct response regimes, most notably an unstudied low-noise ``regime II" where membership signal reverses relative to the traditionally observed high-noise ``regime III".

\partitle{Quantifying model behavior.} For each member and non-member sample, we (1) perturb the sample with additive Gaussian noise, (2) obtain the per-token logits for the original and perturbed sample pair, and (3) compute the per-token divergence between each pair of softmax logit distributions, taking the average of the max-$k$ divergences along the token sequence as the score. This quantifies noise sensitivity of the model regarding member and non-members.
We explore two divergence metrics. The per-token R\'{e}nyi divergence $D^{(\alpha)}_t$ is defined as:
\begin{align}
D^{(\alpha)}_t(p_t\|\tilde{p_t}) =\frac{1}{\alpha-1}\,\log \sum_{v \in V} {p_t(v)}^{\alpha}\,{\tilde{p}_t(v)}^{\,1-\alpha}, \alpha>0,\ \alpha\neq 1.
\label{eq:renyi_div}
\end{align}
Here, $p_t$ and $\tilde{p}_t$ denote the $t$-th token distributions obtained by applying softmax to the original and perturbed logits, respectively. Alternatively, we apply R\'{e}nyi normalization directly to the softmax logits, which sharpens the distribution for $\alpha > 1$, and compute the Kullback--Leibler divergence $D^{(KLD)}$ as:
\begin{align}
D_t^{(\mathrm{KLD})}
\left(q_t^{(\alpha)}\middle\|\tilde{q}_t^{(\alpha)}\right)
&=
\sum_{v\in V}
q_t^{(\alpha)}(v)
\log\left(
\frac{q_t^{(\alpha)}(v)}
{\tilde{q}_t^{(\alpha)}(v)}
\right),
\qquad
q_t^{(\alpha)}(v)
=
\frac{p_t(v)^\alpha}
{\sum_{u\in V}p_t(u)^\alpha}.
\label{eq:renyi_norm}
\end{align}
For $\alpha=\infty$, this normalization is computed as
\begin{align}
q_t^{(\infty)}(v)
&=
\begin{cases}
1/|A_t|, & v\in A_t,\\
0, & \text{otherwise},
\end{cases}
\qquad
A_t=\arg\max_{u\in V}p_t(u),
\label{eq:renyi_inf_norm}
\end{align}
with $\tilde q_t^{(\infty)}$ defined analogously from $\tilde p_t$. Before computing the KL divergence, we clamp each entry of both normalized distributions from below at $\varepsilon=10^{-12}$ and renormalize them, preventing undefined values when either distribution contains zeros.
For a pair of sequences with length $T$, the divergence metric provides $T$ token-wise divergences. A sample's divergence score is the average of the max-$k$ token divergences for a given $k \in [0.1, 1.0]$.

\begin{figure}[h]
    \centering
    \includegraphics[width=0.35\linewidth]{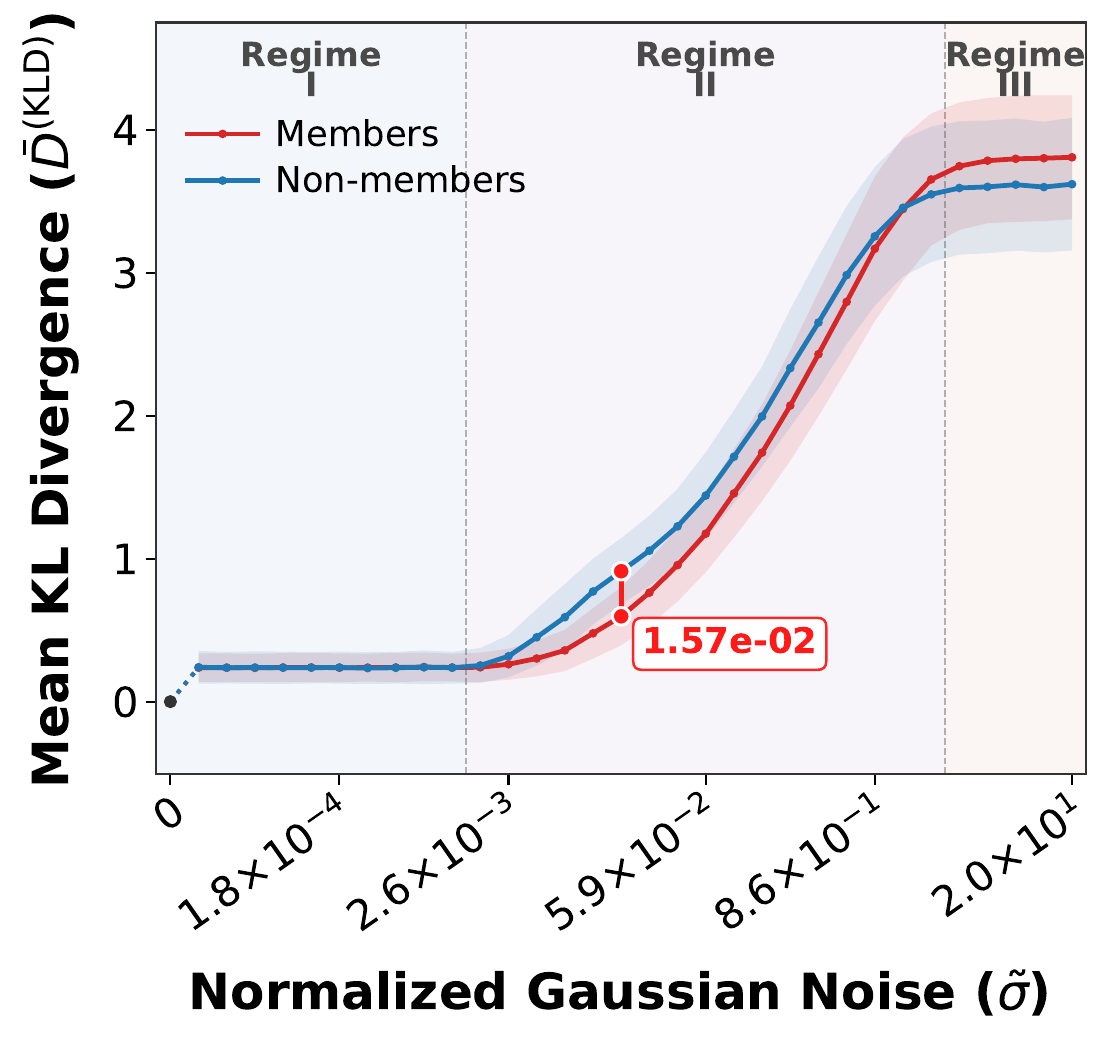}
    \hspace{0.1\linewidth}%
    \includegraphics[width=0.35\linewidth]{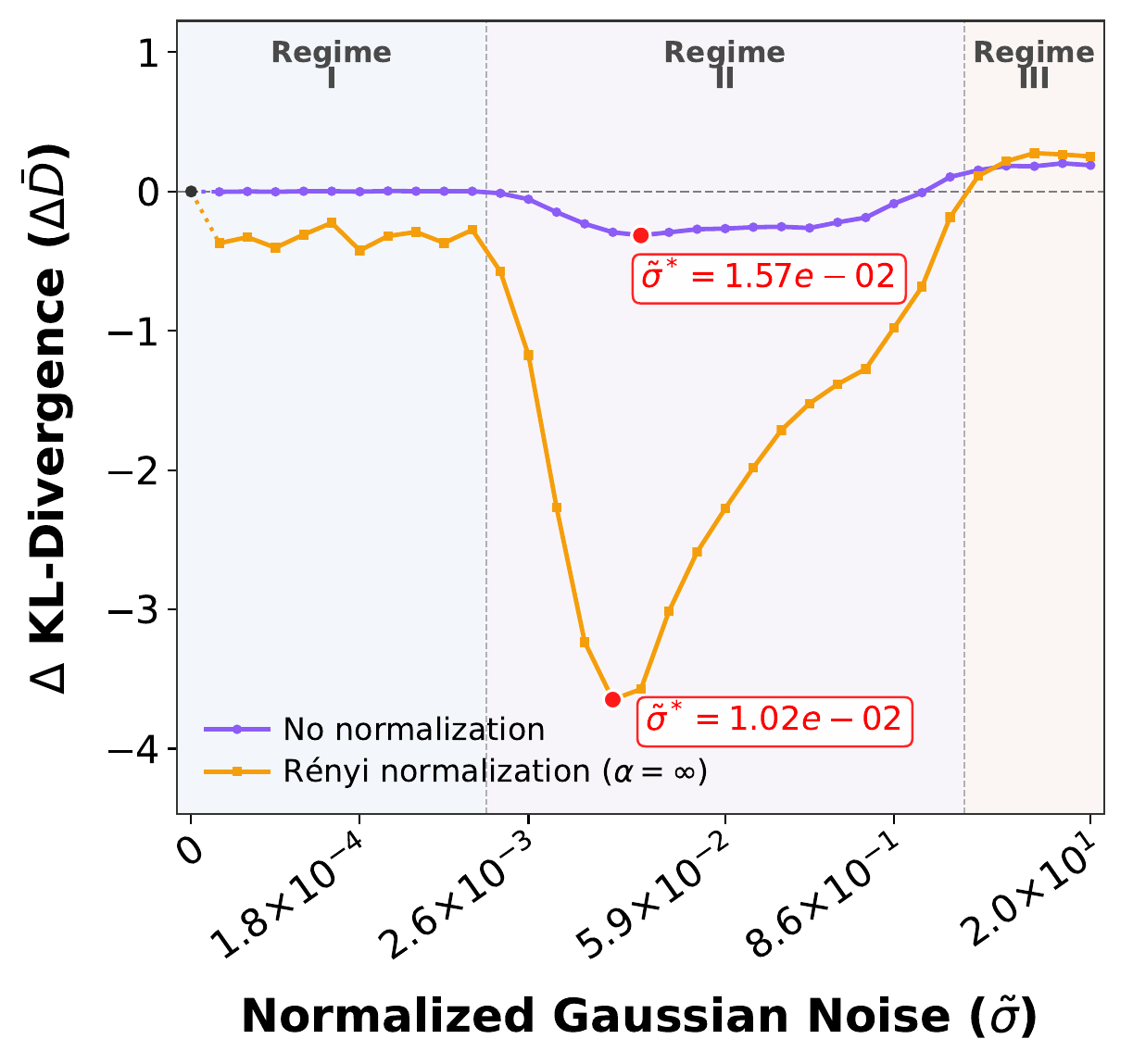}
    \caption{Perturbation sensitivity and membership divergence across normalized Gaussian noise on a Flickr pre-training target set against LLaVA v1.5--7B model with k=0.9. \textbf{Left:} Mean unnormalized KL divergence for members and non-members, with shaded bands denoting one standard deviation. Their similar response in Regime~I produces a near-zero divergence gap, whereas members exhibit greater divergence in Regime~III. \textbf{Right:} The member--non-member gap $\Delta\bar{D}=\bar{D}_m-\bar{D}_n$ with and without R\'enyi normalization ($\alpha=\infty$). R\'enyi normalization amplifies the negative gap in Regime~II. Red markers indicate the noise level minimizing $\Delta\bar{D}$ for each metric.
    }
    \label{fig:kld_dynamics}
\end{figure}

\partitle{Analyzing model sensitivity.} 
For each perturbation magnitude $\tilde{\sigma}$, we compute the mean divergence scores $\bar{D}_m(\tilde{\sigma})$ and $\bar{D}_n(\tilde{\sigma})$ over known member and non-member sets and define their gap as $\Delta \bar{D}(\tilde{\sigma})=\bar{D}_m(\tilde{\sigma})-\bar{D}_n(\tilde{\sigma})$ (\autoref{fig:kld_dynamics}). As $\tilde{\sigma}$ increases, we observe three distinct regimes: (i) a near-zero plateau ($\Delta \bar{D}\approx0$); (ii) a pronounced negative dip under small perturbations ($\Delta \bar{D}<0$), indicating greater divergence for non-members; and (iii) a sharp rise to the conventionally reported behavior under strong perturbations, where members exhibit greater divergence ($\Delta \bar{D}>0$).

\partitle{Regime I: $\boldsymbol{\Delta \bar{D} \approx 0}$.}
At zero noise, both groups exhibit the expected zero divergence, with $\bar{D}_m=\bar{D}_n=0$.
Without normalization, microscopic perturbations induce small, nearly identical increases in $\bar{D}_m$ and $\bar{D}_n$ (\autoref{fig:kld_dynamics}), whereas aggressive normalization reveals only weak separation between the two groups. Given members and non-members respond similarly, Regime I provides little membership signal and is unlikely to reflect memorization of specific training examples.

\partitle{Regime II: $\boldsymbol{\Delta\bar{D}<0}$.}
Regime~II captures our central empirical finding: for some model--dataset pairs, small nonzero perturbations produce greater divergence for non-members than for members, similar to findings of label-only and adversarial-perturbation MIAs against non-generative image classifiers~\cite{choquettechoo2021labelonly,zhang2022evaluating}. This finding contradicts the behavior conventionally observed from VLLMs where members are less robust (member divergences is higher).
Previously, this was not found since existing works have not explored full noise sweep and neglected this low-noise regime.
In fact, we find many cases where this regime offers strong member and non-member separability (bigger gap) and hence we leverage it as a membership signal in the subsequent section.

\partitle{Regime III: $\boldsymbol{\Delta\bar{D}>0}$.}
As the perturbation enters the higher-noise regime, $\bar{D}_m$ crosses over $\bar{D}_n$, returning to the conventionally understood LLM and VLLM behavior~\cite{liu2021encodermi}. Simultaneously, the across-sample standard deviation of the divergence scores increases rapidly for both classes, amplifying variability alongside the mean divergence gap. Beyond the crossover point, perturbed member images become sufficiently dissimilar to the training samples, allowing the model to detect larger distributional shifts. By contrast, we attribute the smaller non-member shifts to
a lack of strong memorized reference in the model's weights, 
ultimately producing a lower asymptotic $D_n$.

\partitle{Leveraging Regime II.} 
Regime~II provides a more discriminative membership signal than its high-noise counterpart. Moreover, we can leverage aggressive normalization to amplify the membership signal, as demonstrated by the substantial negative gap in \autoref{fig:kld_dynamics}. %
We attribute this result to two complementary %
factors: signal strength and signal clarity. First, in regime~II, %
member outputs remain locally anchored while non-member outputs shift away, often producing larger separation than regime~III.
Second, regime~III
has bigger across-sample divergence. Thus, although strong perturbations can enlarge the mean response, they can simultaneously inject variability that obscures the membership signal. Although our results in \autoref{sec:6_experiments} show that regime~II often provides a stronger membership signal, it is not universally optimal: in some settings, regime~III yields better separation. Our proposed attack therefore can leverage both regimes effectively via calibration.

%% file: sec/5_attack_methodology.tex
\section{Membership Inference Attack}
\label{sec:5_attack_methodology}

\begin{figure}
    \centering
    \includegraphics[width=1\linewidth]{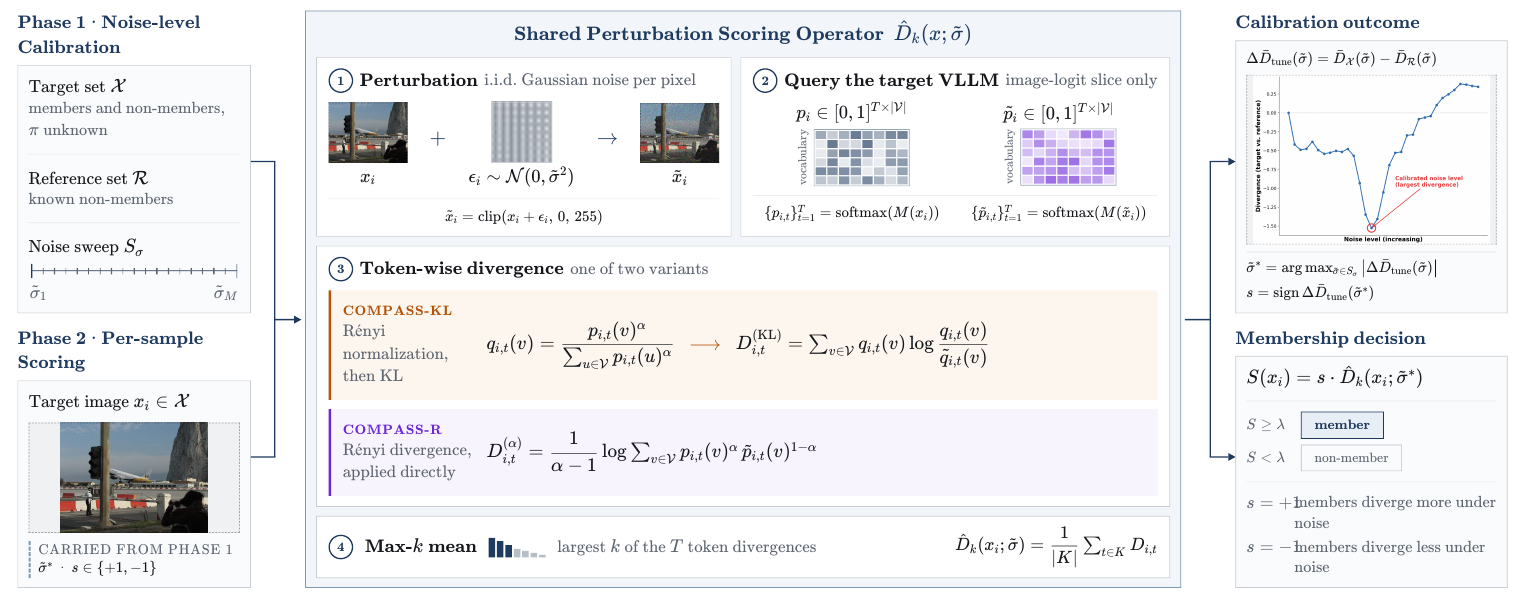}
    \caption{COMPASS attack structure. In the tuning phase, COMPASS compares the divergence responses of the unlabeled target set and a known non-member reference set across candidate noise levels to select the perturbation magnitude and score direction. In the scoring phase, it applies the selected perturbation to each target image, aggregates the largest token-wise divergences, and orients the result to produce a membership score.}
    \label{fig:attack_structure}
\end{figure}

We propose \textbf{COMPASS}, an adaptive membership inference attack that calibrates both the perturbation magnitude and membership-score direction.
Its controlled perturbation-based framework
sweeps a finely resolved range of Gaussian noise magnitudes to identify the strongest membership separation.
This allows dynamic utilization of regime~II or III to find 
the most discriminative perturbation. %
To further refine the signal, COMPASS expands on the general $\Delta D$ score outlined in our perturbation analysis, averaging over the max-$k$ divergence measures to focus the score on the most discriminative features.  
\autoref{fig:attack_structure} summarizes the COMPASS pipeline; we next formalize its perturbation, divergence scoring. We further formalize the calibration procedures in \autoref{app:noise_level_tuning}.

\partitle{Attack structure.}
For each target image
$\mathbf{x}_{i}\in[0,255]^{H\times W\times C}$, we independently sample additive Gaussian noise at each pixel and channel and construct the perturbed image as
\begin{align}
(\boldsymbol{\epsilon}_{i})_{h,w,c}
&\overset{\mathrm{i.i.d.}}{\sim}
\mathcal{N}\!\left(0,(\tilde{\sigma}^{*})^{2}\right),
&
\tilde{\mathbf{x}}_{i}
&=
\operatorname{clip}\!\left(
\mathbf{x}_{i}+\boldsymbol{\epsilon}_{i},\,0,\,255
\right).
\end{align}
where $\tilde{\sigma}^{*}$ denotes the perturbation magnitude selected by the noise-level search as described in \autoref{app:noise_level_tuning}. We then feed the original and perturbed images into the target model to obtain their corresponding image-token probability sequences:
\begin{align}
\{p_{i,t}\}_{t=1}^{T}
&=
\operatorname{softmax}\!\left(
\mathcal{M}(\mathbf{x}_{i})
\right),
&
\{\tilde{p}_{i,t}\}_{t=1}^{T}
&=
\operatorname{softmax}\!\left(
\mathcal{M}(\tilde{\mathbf{x}}_{i})
\right).
\end{align}
Following the perturbation analysis, we compute the divergence between each pair of token-wise distributions using either (i) the R\'enyi divergence $D_{t}^{(\alpha)}$ in \autoref{eq:renyi_div}, yielding COMPASS-R, or (ii) the KL divergence $D_{t}^{(\mathrm{KLD})}$ after R\'enyi normalization in \autoref{eq:renyi_norm}, yielding COMPASS-KL. For COMPASS-KL, we evaluate normalization orders $\alpha\in\{0.5,1.0,2.0,\infty\}$, where $\alpha=\infty$ denotes the limiting max-dominant normalization. For COMPASS-R, we evaluate R\'enyi orders $\alpha\in\{0.25,0.5,2.0,4.0\}$. To obtain a single sample-level membership score while emphasizing the strongest token-wise contributions, we average the top-$k$ divergence values across the sequence. Finally, we orient the score according to the set-specific tuning outcome: if the target set exhibits lower mean divergence than the known non-member set, we negate the aggregated score so that larger values consistently indicate membership.

%% file: sec/6_experiments.tex
\section{Experiments}
\label{sec:6_experiments}
\subsection{Experimental Settings}

\partitle{Target Models.}
We evaluate our attack on LLaVA v1.5--7B~\cite{liu2023visual} and MiniGPT-4~\cite{zhu2023minigpt}, two open-source VLLMs for which substantial information about the underlying training data is publicly available. To broaden our evaluation across model scale and domain, we additionally evaluate the larger medical-domain HuluMed-32B model~\cite{jiang2025hulumed}.

\partitle{Datasets.}
To evaluate membership inference over \textbf{pre-training} data, we use the Flickr benchmark introduced by Li~\textit{et al.}~\cite{li2024membership}, which contains 300 members and 300 non-members for both LLaVA and MiniGPT-4. We randomly sample 150 images from each group to construct each target set. For \textbf{instruction fine-tuning}, we use COCO-2017 images included in LLaVA's disclosed instruction-tuning data~\cite{lin2014microsoft}. To reduce the likelihood that the presumed non-members appeared during pre-training, we draw them from recent Flickr images. Because MiniGPT-4 does not disclose its instruction-tuning data, we restrict this evaluation to LLaVA. For both settings, the tuning reference set is sampled from ShareGPT-4o~\cite{chen2025sharegpt}. For HuluMed-32B, we evaluate COMPASS on endoscopy medical images, pairing
PubMedVision members~\cite{chen2024pubmedvision} with EndoBench non-members~\cite{liu2025endobench} using 
GastroHUN~\cite{bravo2025gastrohun} as the reference set.
Motivated by blind-baseline analyses showing that source differences can confound MIA evaluation~\cite{das2025blind,zhu2025revisiting}, we apply the filtering procedure described in \autoref{app:blind_classifier_filtering} to both self-constructed medical datasets to remove potential bias.

\partitle{Baselines.} We compare our attack with state-of-the-art MIAs on VLLMs. We use (1) \textbf{Aug-KL}~\cite{liu2021encodermi}, which compares the KLD of logits from the original and augmented images; (2) \textbf{Min-$k$\%}~\cite{shi2023detecting}, which uses the average of the min-$k$ token probabilities as the membership signal; and (3) \textbf{Max-R\'enyi-$\alpha$}~\cite{li2024membership}, which uses the average of the max-$k$\% R\'enyi entropies of the output logits. We report only $\alpha=0.5$, their best-performing version.

\subsection{Image MIA on LLaVA \& MiniGPT-4}
\label{subsec:LLaVA Results}

\begin{table*}[h]
\centering
\scriptsize
\setlength{\tabcolsep}{2.5pt}
\renewcommand{\arraystretch}{1.08}

\resizebox{\textwidth}{!}{%
\begin{tabular}{@{}lccc@{\hspace{3pt}}ccc@{\hspace{3pt}}ccc@{}}
\toprule[0.8pt]

& \multicolumn{6}{c}{\textbf{LLaVA v1.5--7B}}
& \multicolumn{3}{c}{\textbf{MiniGPT-4}} \\
\cmidrule(lr){2-7}
\cmidrule(lr){8-10}

& \multicolumn{3}{c}{\textbf{Flickr}}
& \multicolumn{3}{c}{\textbf{COCO-2017}}
& \multicolumn{3}{c}{\textbf{Flickr}} \\
\cmidrule(lr){2-4}
\cmidrule(lr){5-7}
\cmidrule(lr){8-10}

\textbf{Metric}
& $k=0.1$ & $k=0.5$ & $k=0.9$
& $k=0.1$ & $k=0.5$ & $k=0.9$
& $k=0.1$ & $k=0.5$ & $k=0.9$ \\

\midrule[0.8pt]

\multicolumn{10}{l}{\textbf{Baselines}} \\
\cmidrule(lr){1-10}

Aug-KL
& $-$
& $-$
& $0.71 \pm 0.04$
& $-$
& $-$
& $0.69 \pm 0.01$
& $-$
& $-$
& $0.63 \pm 0.01$ \\

Min-$k$\%
& $0.58 \pm 0.02$
& $0.66 \pm 0.05$
& $0.67 \pm 0.05$
& $0.60 \pm 0.01$
& $0.64 \pm 0.03$
& $0.65 \pm 0.02$
& $0.51 \pm 0.04$
& $0.59 \pm 0.02$
& $0.64 \pm 0.02$ \\

Max-R\'enyi$_{0.5\%}$
& $0.64 \pm 0.02$
& $0.72 \pm 0.03$
& $0.70 \pm 0.03$
& $0.69 \pm 0.04$
& $0.74 \pm 0.03$
& $0.72 \pm 0.03$
& $\mathbf{0.78 \pm 0.02}$
& $0.73 \pm 0.01$
& $0.69 \pm 0.01$ \\

\midrule

\multicolumn{10}{l}{\textbf{COMPASS-KL (Ours)}} \\
\cmidrule(lr){1-10}

$\alpha=0.5$
& $0.60^{*} \pm 0.02$
& $0.70^{*} \pm 0.03$
& $0.70^{*} \pm 0.04$
& $0.70^{*} \pm 0.03$
& $0.80^{*} \pm 0.03$
& $0.81^{*} \pm 0.04$
& $0.58^{*} \pm 0.04$
& $0.69^{*} \pm 0.03$
& $0.72^{*} \pm 0.02$ \\

$\alpha=1$
& $0.62^{*} \pm 0.02$
& $0.70^{*} \pm 0.03$
& $0.70^{*} \pm 0.03$
& $0.70^{*} \pm 0.03$
& $0.80^{*} \pm 0.04$
& $0.81^{*} \pm 0.04$
& $0.66^{*} \pm 0.02$
& $0.73^{*} \pm 0.01$
& $0.75^{*} \pm 0.01$ \\

$\alpha=2$
& $\mathbf{0.65^{*} \pm 0.03}$
& $0.71^{*} \pm 0.03$
& $0.72^{*} \pm 0.03$
& $\mathbf{0.74^{*} \pm 0.02}$
& $0.81^{*} \pm 0.05$
& $0.81^{*} \pm 0.04$
& ${0.69^{*} \pm 0.03}$
& $0.72^{*} \pm 0.02$
& $0.75^{*} \pm 0.01$ \\

$\alpha=\infty$
& $0.50^{*} \pm 0.00$
& $\mathbf{0.78^{*} \pm 0.02}$
& $\mathbf{0.78^{*} \pm 0.02}$
& $0.50^{*} \pm 0.00$
& $\mathbf{0.88^{*} \pm 0.04}$
& $\mathbf{0.88^{*} \pm 0.04}$
& $0.53^{*} \pm 0.01$
& ${0.75^{*} \pm 0.02}$
& ${0.77^{*} \pm 0.02}$ \\

\midrule

\multicolumn{10}{l}{\textbf{COMPASS-R (Ours)}} \\
\cmidrule(lr){1-10}

$\alpha=0.25$
& $0.60^{*} \pm 0.03$
& $0.69^{*} \pm 0.02$
& $0.70^{*} \pm 0.02$
& $0.71^{*} \pm 0.03$
& $0.80^{*} \pm 0.04$
& $0.81^{*} \pm 0.04$
& $0.63^{*} \pm 0.03$
& $0.72^{*} \pm 0.02$
& $0.74^{*} \pm 0.01$ \\

$\alpha=0.5$
& $0.59^{*} \pm 0.03$
& $0.69^{*} \pm 0.02$
& $0.69^{*} \pm 0.02$
& $0.71^{*} \pm 0.03$
& $0.80^{*} \pm 0.04$
& $0.81^{*} \pm 0.04$
& $0.64^{*} \pm 0.02$
& $0.71^{*} \pm 0.01$
& $0.74^{*} \pm 0.01$ \\

$\alpha=2$
& ${0.64^{*} \pm 0.02}$
& $0.72^{*} \pm 0.03$
& $0.73^{*} \pm 0.03$
& ${0.73^{*} \pm 0.03}$
& $0.82^{*} \pm 0.04$
& $0.83^{*} \pm 0.04$
& ${0.68^{*} \pm 0.01}$
& $0.76^{*} \pm 0.01$
& $0.79^{*} \pm 0.01$ \\

$\alpha=4$
& ${0.64^{*} \pm 0.02}$
& ${0.73^{*} \pm 0.03}$
& ${0.74^{*} \pm 0.03}$
& $0.72^{*} \pm 0.02$
& ${0.84^{*} \pm 0.03}$
& ${0.85^{*} \pm 0.04}$
& ${0.68^{*} \pm 0.01}$
& $\mathbf{0.77^{*} \pm 0.01}$
& $\mathbf{0.81^{*} \pm 0.01}$ \\

\bottomrule[0.8pt]
\end{tabular}%
}

\caption{
Average AUC over three independently sampled target sets for LLaVA v1.5--7B on Flickr and COCO-2017, and MiniGPT-4 on Flickr.
An asterisk indicates that the COMPASS decided to use region~II for membership signal.
}
\label{tab:llava_minigpt_auc_results}
\end{table*}

\partitle{Attack Performance.}
\autoref{tab:llava_minigpt_auc_results} compares the AUC of COMPASS-KL and COMPASS-R against the considered baselines for LLaVA v1.5--7B and MiniGPT-4. Since Min-$k$\%, Max-R\'enyi, and our proposed attacks operate on a top-$k$ subset of output tokens, we report representative settings $k\in\{0.1,0.5,0.9\}$ and highlight the best-performing configuration for each method. Aug-KL instead computes its membership score over the full output sequence and is therefore reported once under $k=0.9$ for presentation. All results are averaged over three independently sampled target sets, with standard deviation reported across runs.

On the Flickr pre-training data, COMPASS provides strong membership separation across both evaluated VLLMs. For LLaVA, COMPASS-KL achieves a maximum AUC of $0.78\pm0.02$, improving over the strongest baseline at $0.72\pm0.03$, while COMPASS-R reaches $0.74\pm0.03$. We observe comparable performance on MiniGPT-4. Although Max-R\'enyi-$0.5\%$ provides a strong baseline at $0.78\pm0.02$, COMPASS-R achieves the highest overall AUC of $0.81\pm0.01$, while COMPASS-KL reaches $0.77\pm0.02$. In both cases where COMPASS obtains the strongest result, its standard deviation is also no greater than that of the strongest competing baseline, indicating that the improvement persists across independently sampled target sets rather than arising from a single favorable sample.

The improvement is substantially larger on LLaVA's COCO-2017 instruction-tuning data. COMPASS-KL achieves a maximum AUC of $0.88\pm0.04$ and COMPASS-R reaches $0.85\pm0.04$, compared with $0.74\pm0.03$ for the strongest baseline. Although the variation is slightly larger in this setting, the gain in mean AUC is substantially greater. In particular, COMPASS-KL with R\'enyi-$\infty$ normalization achieves $0.88\pm0.04$ at both $k=0.5$ and $k=0.9$. Overall, COMPASS achieves the highest AUC across all three evaluated model--dataset settings while exhibiting comparable, and in several cases lower, variation across target-set samples.

\begin{wrapfigure}{r}{0.40\textwidth}
    \vspace{-1.25\baselineskip}
    \centering
    \setlength{\abovecaptionskip}{2pt}
    \setlength{\belowcaptionskip}{0pt}

    \includegraphics[
        width=\linewidth
    ]{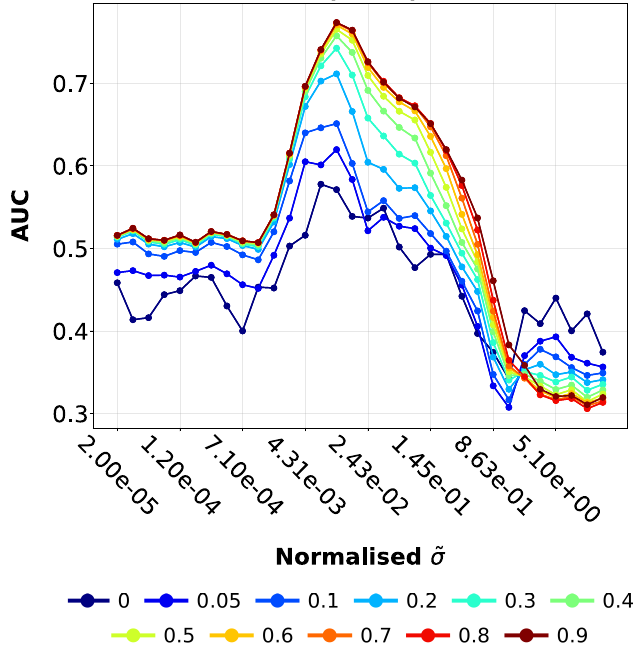}

    \caption{Max-$k$ effect for COMPASS-R on Flickr.}
    \label{fig:max_k_effect}
    \vspace{-\baselineskip}
\end{wrapfigure}

\partitle{Effect of R\'enyi Order.}
Performance generally improves as the R\'enyi order $\alpha$ increases for both COMPASS variants. Larger $\alpha$ places greater emphasis on dominant output probabilities, with R\'enyi-$\infty$ normalization producing the strongest COMPASS-KL results and the largest evaluated order similarly performing best for COMPASS-R. This suggests that 
COMPASS's membership signal is concentrated 
primarily among the model's dominant output
probabilities.

{
\setlength{\columnsep}{7pt}

\partitle{Effect of Max-$\bm{k}$.}
\autoref{fig:max_k_effect} shows that, across noise levels, retaining a large
top-$k$ fraction of token-wise divergence values
($k\in\{0.7,0.8,0.9\}$) consistently yields higher AUCs and a stronger
membership signal in the pre-training setting. We observe the same pattern
in the COCO-2017 instruction-fine-tuning setting shown by \autoref{tab:llava_minigpt_auc_results}. We hypothesize that this
behavior arises from the distributed nature of visual representations: in
the evaluated models, membership-relevant evidence may be spread across many
correlated visual tokens rather than concentrated in only a few. Aggressive
truncation at small $k$ can therefore discard weak but correlated cues,
increasing variance and attenuating member--non-member separation. Retaining
a larger $k$ preserves this distributed evidence, stabilizes the estimate,
and strengthens the resulting separation.

\par
}

\subsection{Medical Image MIA on HuluMed-32B}

\begin{table*}[h]
\centering
\scriptsize
\setlength{\tabcolsep}{2.8pt}
\renewcommand{\arraystretch}{1.12}

\begin{tabular}{@{}lccc@{\hspace{7pt}}ccc@{}}
\toprule[0.8pt]
& \multicolumn{3}{c}{\textbf{Endoscopy (AUC)}}
& \multicolumn{3}{c}{\textbf{Endoscopy (TPR@5\%FPR)}} \\
\cmidrule(lr){2-4}
\cmidrule(lr){5-7}

\textbf{Metric}
& $k=0.7$ & $k=0.8$ & $k=0.9$
& $k=0.7$ & $k=0.8$ & $k=0.9$ \\
\midrule[0.8pt]

\multicolumn{7}{l}{\textbf{Baselines}} \\
\cmidrule(lr){1-7}

Aug-KL
& $-$ & $-$ & $0.42 \pm 0.04$
& $-$ & $-$ & $0.01 \pm 0.00$ \\

Min-$k$\%
& $0.50 \pm 0.00$
& $0.50 \pm 0.00$
& $0.50 \pm 0.00$
& $0.00 \pm 0.00$
& $0.00 \pm 0.00$
& $0.00 \pm 0.00$ \\

Max-R\'enyi$_{0.5\%}$
& $0.50 \pm 0.00$
& $0.50 \pm 0.00$
& $0.50 \pm 0.00$
& $0.00 \pm 0.00$
& $0.00 \pm 0.00$
& $0.00 \pm 0.00$ \\

\midrule
\multicolumn{7}{l}{\textbf{COMPASS-KL (Ours)}} \\
\cmidrule(lr){1-7}

$\alpha = 0.5$
& $0.58 \pm 0.03$
& $0.57 \pm 0.03$
& $0.57 \pm 0.03$
& $0.12 \pm 0.02$
& $0.12 \pm 0.02$
& $0.13 \pm 0.02$ \\

$\alpha = 1.0$
& $0.61^{*} \pm 0.00$
& $0.62^{*} \pm 0.00$
& $0.62^{*} \pm 0.00$
& $0.12^{*} \pm 0.01$
& $0.12^{*} \pm 0.01$
& $0.13^{*} \pm 0.01$ \\

$\alpha = 2.0$
& $0.60^{*} \pm 0.01$
& $0.61^{*} \pm 0.01$
& $0.61^{*} \pm 0.01$
& $0.12^{*} \pm 0.01$
& $0.11^{*} \pm 0.00$
& $0.11^{*} \pm 0.01$ \\

$\alpha = \infty$
& $0.58^{*} \pm 0.03$
& $0.59^{*} \pm 0.02$
& $0.60^{*} \pm 0.02$
& $0.13^{*} \pm 0.02$
& $0.10^{*} \pm 0.04$
& $0.10^{*} \pm 0.04$ \\

\midrule
\multicolumn{7}{l}{\textbf{COMPASS-R (Ours)}} \\
\cmidrule(lr){1-7}

$\alpha = 0.25$
& $\mathbf{0.63^{*} \pm 0.01}$
& $\mathbf{0.63^{*} \pm 0.01}$
& $\mathbf{0.64^{*} \pm 0.01}$
& $0.11^{*} \pm 0.04$
& $0.12^{*} \pm 0.04$
& $0.13^{*} \pm 0.05$ \\

$\alpha = 0.5$
& $\mathbf{0.63^{*} \pm 0.01}$
& $\mathbf{0.63^{*} \pm 0.01}$
& $0.63^{*} \pm 0.01$
& $0.12^{*} \pm 0.03$
& $0.13^{*} \pm 0.04$
& $\mathbf{0.14^{*} \pm 0.05}$ \\

$\alpha = 2.0$
& $0.62^{*} \pm 0.01$
& $0.62^{*} \pm 0.01$
& $0.62^{*} \pm 0.01$
& $\mathbf{0.14^{*} \pm 0.03}$
& $0.14^{*} \pm 0.03$
& $\mathbf{0.14^{*} \pm 0.03}$ \\

$\alpha = 4.0$
& $0.59^{*} \pm 0.03$
& $0.62^{*} \pm 0.01$
& $0.62^{*} \pm 0.02$
& $0.12^{*} \pm 0.03$
& $\mathbf{0.15^{*} \pm 0.02}$
& $0.13^{*} \pm 0.03$ \\

\bottomrule[0.8pt]
\end{tabular}

\caption{
Average AUC and TPR@5\%FPR for the Endoscopy dataset against HuluMed-32B. Following the finding that large Max-$k$ yields stronger performance, we report representative top-$k$ settings $k\in\{0.7,0.8,0.9\}$ evaluated under the same pipeline as LLaVA and Minigpt (\autoref{subsec:LLaVA Results}). Bold indicates the highest reported value in each column, including ties.An asterisk indicates that the COMPASS decided to use region~II for membership signal.
}
\label{tab:hulumed32b_endoscopy_auc_tpr_results}
\end{table*}

\autoref{tab:hulumed32b_endoscopy_auc_tpr_results} reports AUC
for HuluMed-32B on the filtered Endoscopy dataset. Given membership inference is more challenging in this setting, we also report TPR@5\%FPR to assess detection performance at a low false-positive rate. Min-$k$\% and Max-R\'enyi perform at chance-level AUC ($0.50$) and achieve zero TPR@5\%FPR across the reported $k$ values. Aug-KL reaches $0.42\pm0.04$ AUC and $0.01\pm0.00$ TPR@5\%FPR. These results show that the considered baselines provide little membership separation in this setting.

In contrast, COMPASS achieves membership inference, indicating that its perturbation scores retain a membership signal in the harder endoscopy setting. COMPASS-R with $\alpha=0.25$ reaches the highest AUC of $0.64$ at $k=0.9$; with $\alpha=0.5$, it reaches $0.63$ across all three reported $k$ values. COMPASS-KL reaches $0.62$ AUC with $\alpha=1$ at $k=0.8$ and $k=0.9$. The advantage over the baselines is more evident at low false-positive rates. At 5\% FPR, COMPASS-R with $\alpha=4$ achieves the highest TPR of $0.15$ at $k=0.8$, compared with $0.01$ for Aug-KL and $0.00$ for Min-$k$\% and Max-R\'enyi. The best COMPASS-KL configuration reaches $0.13$ TPR at $k=0.9$.

Asterisks in \autoref{tab:llava_minigpt_auc_results} and \autoref{tab:hulumed32b_endoscopy_auc_tpr_results} shows the significance of our calibration. With the reference set, it scans regimes and determines the ideal separation for each dataset, leading to higher AUC.

\subsection{Noise-level Tuning Effectiveness}
\label{sec:Noise-level Tuning Effectiveness}

\begin{figure*}[h]
    \raggedright

    \begin{minipage}[t]{0.53\textwidth}
        \vspace{0pt}
        \raggedright
        \small
        \setlength{\tabcolsep}{4pt}
        \renewcommand{\arraystretch}{1.15}
        \begin{tabular}{@{}lcc@{}}
            \toprule
            Dataset & Optimal $\tilde{\sigma}$ & Estimated $\tilde{\sigma}$ \\
                    & AUC & AUC \\
            \midrule
            LLaVA Flickr      & $0.80\pm0.018$ & $0.78\pm0.02$ \\
            LLaVA COCO-2017   & $0.92\pm0.01$  & $0.88\pm0.04$ \\
            HuluMed Endoscopy & $0.69\pm0.02$ & $0.63\pm0.02$ \\
            \bottomrule
        \end{tabular}

        \vspace{6pt}
        \caption{
            Effectiveness of noise-level tuning.
            \textbf{Top:} AUC using the true optimal noise level and
            the noise level estimated by our tuning procedure.
            \textbf{Right:} Estimated and true optimal noise levels for
            R\'enyi-$\infty$ normalized KL divergence difference. The estimated
            $\tilde{\sigma}$ lies near the true optimum and yields
            performance close to the ideal setting.
        }
        \label{fig:noise-level-tuning}
    \end{minipage}%
    \hspace{0.04\textwidth}%
    \begin{minipage}[t]{0.4\textwidth}
        \vspace{0pt}
        \centering
        \includegraphics[width=\linewidth]{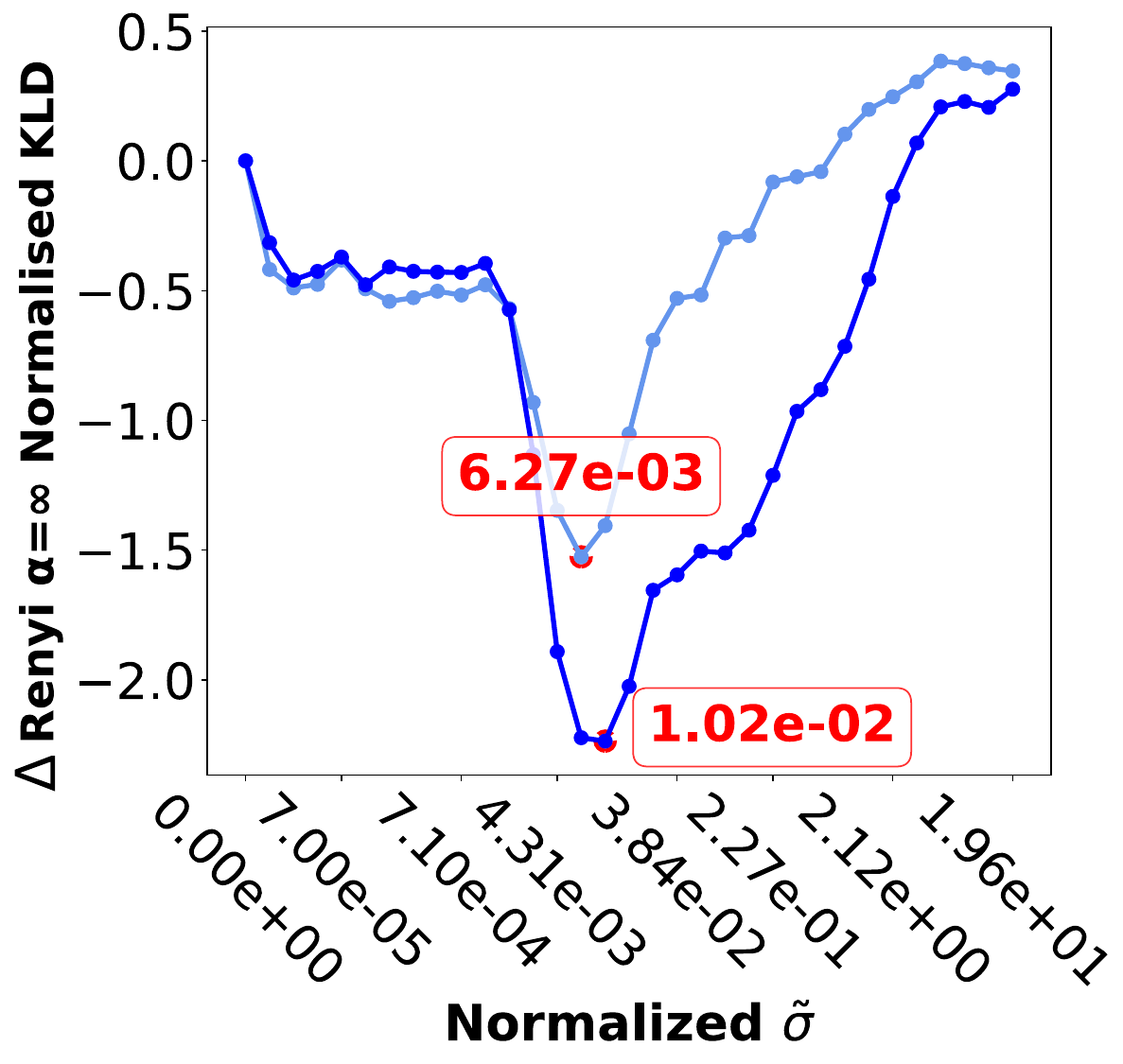}\\[-2pt]
        \includegraphics[width=0.85\linewidth]{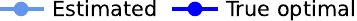}
    \end{minipage}

    \vspace{-4pt}
\end{figure*}

We find that the tuning procedure for the noise magnitude $\tilde{\sigma}^*$ described in \autoref{sec:5_attack_methodology} is a strong estimator of the true optimum. To demonstrate this, we run the attack over the entire range of noise levels and identify the oracle perturbation noise level corresponding to the strongest AUC. \autoref{fig:noise-level-tuning} (Top) compares this ideal AUC with the empirical AUC we obtain, showing that the two are comparable.

\autoref{fig:noise-level-tuning} visually demonstrates the effectiveness of the proposed noise tuning method while highlighting the importance of its adaptive nature.
While the scale of the $\Delta D$ drop is reduced by the member--non-member ratio $\pi$, the estimate still yields a $\tilde{\sigma}$ that lies within the trough of the minimum. Although the $\delta(\tilde{\sigma})$ term introduces a slight deviation from the true optimum $\tilde{\sigma}^*$, the results in \autoref{fig:noise-level-tuning} demonstrate that the estimated $\tilde{\sigma}$ remains a strong differentiating signal for our attack.

%% file: sec/7_conclusion.tex
\section{Conclusion}
\label{sec:7_conclusion}
In this paper, we systematically characterized the membership behavior of diverse VLLMs under image perturbations across a finely resolved range of noise levels, revealing behavior substantially more complex than conventionally assumed. In particular, we found that, for many model--dataset pairs, members exhibit lower divergence than non-members under extremely small perturbations, reversing the conventional membership-signal direction and producing stronger membership separation than the conventionally considered high-noise regime. Building on this observation, we introduced COMPASS-KL and COMPASS-R, two self-calibrating MIA variants that search the full perturbation range to identify both the most discriminative noise level and the corresponding membership-score direction using the target set and a known non-member reference set, without requiring labeled members. Across our experiments, COMPASS consistently outperformed existing baselines, demonstrating the value of adapting perturbation-based MIAs to the noise-response characteristics of each target setting. Future work will investigate stronger theoretical guarantees for noise-level tuning and develop more robust estimators under reference-set distribution shift and varying target-set member proportions.

%% file: sec/8_appendix.tex
\section*{Appendix}

\section*{Contents of the Appendix}
\label{app:Contents of the Appendix}
We provide supporting materials and analyses in the appendix. In \autoref{app:Assets}, we link to the anonymized source code and datasets used in our evaluations. \autoref{app:Experimental Setup} identifies the target models and describes the hardware and runtime requirements. In \autoref{app:algorithms}, we present the COMPASS membership-scoring and noise-level tuning algorithms, followed by a derivation of the tuning gap and its potential error from reference-set differences in \autoref{app:noise_level_tuning}. \autoref{app:blind_classifier_filtering} describes how we filter the endoscopy candidate pools to reduce image-source confounding. Finally, \autoref{app:Additional Results} shows example images across the tested noise levels and the raw divergence curves for each target model, further illustrating the low-noise membership behavior discussed in \autoref{sec:4_the_low_std_region}.

\section{Assets}
\label{app:Assets}
\subsection{Source Code}
Our source code is available in the anonymized GitHub repository:
\url{https://anonymous.4open.science/r/COMPASS_anonymous_github-07A2/readme.md}.
The source code was adapted from Li et al.~\cite{li2024membership}.

\subsubsection{Datasets}
We use three evaluation settings to assess COMPASS. To evaluate membership inference risk for the pre-training data of LLaVA and MiniGPT-4o, we construct target sets of members and non-members from the models' provided Flickr subsets \href{https://huggingface.co/datasets/JaineLi/VL-MIA-image}{https://huggingface.co/datasets/JaineLi/VL-MIA-image} \cite{li2024membership}. To evaluate risk for LLaVA's instruction fine-tuning data, we use the provided COCO-2017 subset \href{https://huggingface.co/datasets/liuhaotian/LLaVA-Instruct-150K/viewer}{https://huggingface.co/datasets/liuhaotian/LLaVA-Instruct-150K/viewer} \cite{li2024membership}. For both settings, we tune the noise level using the disjoint ShareGPT-4o-Image dataset \href{https://huggingface.co/datasets/FreedomIntelligence/ShareGPT-4o-Image}{https://huggingface.co/datasets/FreedomIntelligence/ShareGPT-4o-Image} \cite{chen2025sharegpt}.

To evaluate the larger Hulu-Med-32B model, we construct target sets using the filtering procedure described in \autoref{app:blind_classifier_filtering}. We use PubMedVision endoscopy images \href{https://huggingface.co/datasets/FreedomIntelligence/PubMedVision}{https://huggingface.co/datasets/FreedomIntelligence/PubMedVision} \cite{chen2024pubmedvision} as members and images from EndoBench \href{https://huggingface.co/datasets/Saint-lsy/EndoBench}{https://huggingface.co/datasets/Saint-lsy/EndoBench} \cite{liu2025endobench} as non-members. We tune the perturbation noise level using the GastroHUN dataset \href{https://huggingface.co/datasets/doantrongthai/GastroHun/tree/main}{https://huggingface.co/datasets/doantrongthai/GastroHun/tree/main} \cite{bravo2025gastrohun}.

\section{Experimental Setup}
\label{app:Experimental Setup}
\subsection{Models}
We evaluate our membership inference attacks on three vision–language models, each of which has been pre-trained and subsequently instruction fine-tuned. The first target model is LLaVA-v1.5-7B, released by Liu et al.~\cite{liu2023visual} and publicly available at
\href{https://huggingface.co/liuhaotian/llava-v1.5-7b}{https://huggingface.co/liuhaotian/llava-v1.5-7b}. The second target model is MiniGPT-4, proposed by Zhu et al.~\cite{zhu2023minigpt} and available at
\href{https://github.com/Vision-CAIR/MiniGPT-4}{https://github.com/Vision-CAIR/MiniGPT-4}. The third target model is Hulu-Med-32B released by \cite{jiang2025hulumed} and available at \href{https://github.com/ZJUI-AI4H/Hulu-Med}{https://github.com/ZJUI-AI4H/Hulu-Med}.

\subsection{Processor Specifications}
All experiments are run on a single NVIDIA H200 GPU with 140\,GB of GPU memory. For a combined target and tuning set of 600 samples, collecting all metric values across 32 noise levels requires approximately 12 hours for LLaVA-v1.5-7B and 4 hours for MiniGPT-4. The larger Hulu-Med-32B model requires approximately 20 hours.

\section{COMPASS Algorithms}
\label{app:algorithms}

\subsection{COMPASS Membership Scoring Algorithm}
Following the pipeline described in \autoref{sec:5_attack_methodology}, algorithm~\ref{alg:compass-score} details how COMPASS computes a membership score for a target image. It compares the image-token distributions obtained from the original and perturbed images, computes token-wise divergences using either COMPASS-KL or COMPASS-R, and averages the largest top-$k$ fraction. The noise level and score direction are supplied by the tuning procedure, and the algorithm returns the resulting oriented membership score.
\begin{algorithm}[H]
\caption{COMPASS-KL \& COMPASS-R: Perturbation-Based Membership Scoring}
\label{alg:compass-score}
\begin{algorithmic}[1]
\Require Target image $X\!\in\![0,255]^{H\times W\times C}$,
         black-box model $\mathcal{M}$,
         selected noise std $\tilde{\sigma}^{*}$,
         selected score direction $s\!\in\!\{-1,+1\}$,
         top-$k$ fraction $k\!\in\!(0,1]$,
         variant $v\!\in\!\{\mathrm{KL},\mathrm{R}\}$,
         R\'enyi order $\alpha$
\Ensure Oriented membership score $S(X)$

\State Sample $\epsilon_{h,w,c} \overset{\mathrm{i.i.d.}}{\sim}
       \mathcal{N}(0,(\tilde{\sigma}^{*})^{2})$
\State $\tilde{X} \gets
       \mathrm{clip}\!\left(X+\epsilon,\ 0,255\right)$

\State $\{p_t\}_{t=1}^{T}
       \gets \mathrm{softmax}\!\big(\mathcal{M}(X)\big)$
       \Comment{original image-logit distributions}
\State $\{\tilde{p}_t\}_{t=1}^{T}
       \gets \mathrm{softmax}\!\big(\mathcal{M}(\tilde{X})\big)$
       \Comment{perturbed image-logit distributions}

\If{$v=\mathrm{KL}$} \Comment{COMPASS-KL}
    \For{$t=1$ to $T$}
        \State $q_t(u)
        \gets
        \dfrac{p_t(u)^{\alpha}}
        {\sum_{w\in\mathcal{V}}p_t(w)^{\alpha}}$
        \State $\tilde{q}_t(u)
        \gets
        \dfrac{\tilde{p}_t(u)^{\alpha}}
        {\sum_{w\in\mathcal{V}}\tilde{p}_t(w)^{\alpha}}$
        \State $d_t
        \gets
        D_{\mathrm{KL}}\!\left(q_t\,\|\,\tilde{q}_t\right)$
    \EndFor
\Else \Comment{COMPASS-R}
    \For{$t=1$ to $T$}
        \State $d_t
        \gets
        D_{\alpha}\!\left(p_t\,\|\,\tilde{p}_t\right)$
        \Statex \hspace{\algorithmicindent}
        $\displaystyle
        \phantom{d_t}
        =
        \frac{1}{\alpha-1}
        \log\!\sum_{u\in\mathcal{V}}
        p_t(u)^{\alpha}
        \tilde{p}_t(u)^{1-\alpha}$
    \EndFor
\EndIf

\State $\mathcal{I}_k
       \gets$ indices of the largest
       $\lceil kT\rceil$ values in $\{d_t\}_{t=1}^{T}$
\State $D(X;\tilde{\sigma}^{*})
       \gets
       \dfrac{1}{|\mathcal{I}_k|}
       \sum_{t\in\mathcal{I}_k} d_t$
       \Comment{max-$k$ aggregation}
\State $S(X) \gets s\,D(X;\tilde{\sigma}^{*})$
       \Comment{larger scores favor membership}

\State \Return $S(X)$
\end{algorithmic}
\end{algorithm}

\subsection{COMPASS Noise-Level Tuning}
Algorithm~\ref{alg:compass-cal} selects the perturbation magnitude and score direction used by Algorithm~\ref{alg:compass-score}. For each candidate noise level, it compares the mean raw divergence of the unlabeled target set with that of a known non-member reference set. It selects the noise level with the largest absolute difference and uses the sign of that difference to orient subsequent membership scores.

\begin{algorithm}[H]
\caption{COMPASS Reference-Guided Noise and Score-Orientation Calibration}
\label{alg:compass-cal}
\begin{algorithmic}[1]
\Require Target set $\mathcal{X}$ with unknown membership,
         known non-member reference set $\mathcal{R}$,
         candidate noise grid $\Sigma$,
         black-box model $\mathcal{M}$,
         top-$k$ fraction $k$,
         variant $v\!\in\!\{\mathrm{KL},\mathrm{R}\}$,
         R\'enyi order $\alpha$
\Ensure Selected noise level $\tilde{\sigma}^{*}$,
        score direction $s\!\in\!\{-1,+1\}$

\For{each $\tilde{\sigma}\in\Sigma$}
    \State Compute raw $D(X;\tilde{\sigma})$ using
           the divergence and aggregation steps of
           Algorithm~\ref{alg:compass-score}
           for all $X\in\mathcal{X}$
    \State Compute raw $D(R;\tilde{\sigma})$ using
           the same steps for all $R\in\mathcal{R}$

    \State $\bar{D}_{\mathcal{X}}(\tilde{\sigma})
           \gets
           \dfrac{1}{|\mathcal{X}|}
           \sum_{X\in\mathcal{X}}D(X;\tilde{\sigma})$

    \State $\bar{D}_{\mathcal{R}}(\tilde{\sigma})
           \gets
           \dfrac{1}{|\mathcal{R}|}
           \sum_{R\in\mathcal{R}}D(R;\tilde{\sigma})$

    \State $\Delta\bar{D}_{\mathrm{tune}}(\tilde{\sigma})
           \gets
           \bar{D}_{\mathcal{X}}(\tilde{\sigma})
           -
           \bar{D}_{\mathcal{R}}(\tilde{\sigma})$
\EndFor

\State $\displaystyle
       \tilde{\sigma}^{*}
       \gets
       \arg\max_{\tilde{\sigma}\in\Sigma}
       \left|
       \Delta\bar{D}_{\mathrm{tune}}(\tilde{\sigma})
       \right|$

\If{$\Delta\bar{D}_{\mathrm{tune}}
      (\tilde{\sigma}^{*}) \geq 0$}
    \State $s\gets +1$
    \Comment{higher divergence favors membership}
\Else
    \State $s\gets -1$
    \Comment{lower divergence favors membership}
\EndIf

\State \Return $(\tilde{\sigma}^{*},s)$
\end{algorithmic}
\end{algorithm}

\section{Noise Level Tuning}
\label{app:noise_level_tuning}
\subsection{Noise Level Tuning Procedure}
While \autoref{sec:4_the_low_std_region} shows that each target model--dataset pair has an optimal noise level for membership separation, a practical tuning procedure must operate without membership labels and determine the appropriate score direction for the selected regime.
We therefore use the unlabeled target set as a proxy member set and compare its aggregate divergence response with that of a known non-member reference set, which may be readily obtained through synthetic generation or temporal cutoffs. We reason that the member component remains present in the aggregate response of the target set. Assuming that target and reference non-members exhibit comparable divergence behavior, subtracting their set-level averages approximately cancels the baseline non-member response. Provided that the residual mismatch remains small across the noise sweep, the resulting signal is attenuated by the presence of target non-members but retains the direction and general shape of the underlying membership-separation curve. COMPASS therefore selects the noise level with the largest absolute discrepancy and uses its sign to orient the membership score. Because tuning is performed directly on the target set that is subsequently evaluated, the selected operating point is tailored to the target samples without requiring labeled members. We formalize this tuning signal in \autoref{app:tuning_error}, deriving an expression that captures the effect of distributional mismatch between the target and reference sets. In \autoref{sec:Noise-level Tuning Effectiveness}, we further evaluate the tuning procedure empirically across varying degrees of mismatch, demonstrating its effectiveness even under substantial distributional shift.

\subsection{Noise Level Tuning Error Expression}
\label{app:tuning_error}
COMPASS selects its noise level and score direction by comparing the mean raw divergence of the unlabeled target set $\mathcal{X}$ with that of a known non-member reference set $\mathcal{R}$. As $\mathcal{X}$ contains both members and non-members, we intuit that its mean response retains a membership-dependent component. At each candidate noise level, the observable tuning gap is
\begin{align}
\Delta\bar{D}_{\mathrm{tune}}(\tilde{\sigma})
&=
\bar{D}_{\mathcal{X}}(\tilde{\sigma})
-
\bar{D}_{\mathcal{R}}(\tilde{\sigma}).
\label{eq:divergence_estimator}
\end{align}
Hence, while this gap it is not itself a score for any individual image, it reflects the general shape and score direction that separates the target set members and non-members. The derivation below separates the contribution of membership from the discrepancy between target and reference non-members.

Formally, let $\pi\in[0,1]$ denote the unknown proportion of members in the target set $\mathcal{X}$, and let $\bar{D}_{m}(\tilde{\sigma})$ and $\bar{D}_{n,\mathcal{X}}(\tilde{\sigma})$ denote the expected divergence of its member and non-member components, respectively. The expected target-set divergence can then be expressed as
\begin{align}
\bar{D}_{\mathcal{X}}(\tilde{\sigma})
&=
\pi\bar{D}_{m}(\tilde{\sigma})
+
(1-\pi)\bar{D}_{n,\mathcal{X}}(\tilde{\sigma}).
\label{eq:target_mixture}
\end{align}

We define the discrepancy between the target non-members and the known non-member reference set $\mathcal{R}$ as
\begin{align}
\delta_{\mathcal{R}}(\tilde{\sigma})
&=
\bar{D}_{n,\mathcal{X}}(\tilde{\sigma})
-
\bar{D}_{\mathcal{R}}(\tilde{\sigma}).
\label{eq:reference_discrepancy}
\end{align}
Defining the desired signal as $\Delta\bar{D}(\tilde{\sigma})=\bar{D}_{m}(\tilde{\sigma})-\bar{D}_{n,\mathcal{X}}(\tilde{\sigma})$, the observable tuning gap can then be expressed as:
\begin{align}
\Delta\bar{D}_{\mathrm{tune}}(\tilde{\sigma})
&=
\bar{D}_{\mathcal{X}}(\tilde{\sigma})
-
\bar{D}_{\mathcal{R}}(\tilde{\sigma})
\nonumber \\
&=
\pi\bar{D}_{m}(\tilde{\sigma})
+
(1-\pi)\bar{D}_{n,\mathcal{X}}(\tilde{\sigma})
-
\bar{D}_{\mathcal{R}}(\tilde{\sigma})
\nonumber \\
&=
\pi\left[
\bar{D}_{m}(\tilde{\sigma})
-
\bar{D}_{n,\mathcal{X}}(\tilde{\sigma})
\right]
+
\delta_{\mathcal{R}}(\tilde{\sigma})
\nonumber \\
&=
\pi\Delta\bar{D}(\tilde{\sigma})
+
\delta_{\mathcal{R}}(\tilde{\sigma}).
\label{eq:tuning_gap}
\end{align}
Thus, the observable gap comprises the desired membership signal, attenuated by the unknown member proportion $\pi$, and a reference-set discrepancy term $\delta_{\mathcal{R}}(\tilde{\sigma})$. If, around the dominant extrema of the noise sweep,
$\left|\delta_{\mathcal{R}}(\tilde{\sigma})\right| \ll \pi\left|\Delta\bar{D}(\tilde{\sigma})\right|,$
then the observable tuning gap approximately preserves the direction and relative magnitude of the underlying membership gap: $\Delta\bar{D}_{\mathrm{tune}}(\tilde{\sigma}) \approx \pi\Delta\bar{D}(\tilde{\sigma})$.

We empirically verify this condition using reference sets spanning varying degrees of distributional shift from the target non-members (\autoref{fig:noise-level-tuning}). Across all evaluated model--dataset and reference-set configurations, the observable tuning gap retains the direction and general noise-response pattern of the true membership gap, including when the reference set is substantially distributionally shifted. These results demonstrate that, in our evaluated settings, the membership component remains dominant over the reference-set discrepancy near the most discriminative regions of the noise sweep.

Because $\pi$ is fixed for a given target set and independent of $\tilde{\sigma}$, it uniformly scales the underlying membership gap. For any $\pi>0$, this scaling alone changes neither the direction of $\Delta\bar{D}(\tilde{\sigma})$ nor the noise level maximizing its absolute magnitude. The value of $\pi$ therefore need not be known by the attacker. Moreover, a larger member proportion or a stronger underlying membership gap increases the tolerance of the tuning procedure to the reference-set discrepancy $\delta_{\mathcal{R}}(\tilde{\sigma})$. In practice, our experimental target sets fix $\pi=0.5$, and we demonstrate that our tuning method remains effective in this setting. Accordingly, COMPASS selects the perturbation magnitude
\begin{align}
\tilde{\sigma}^{*}
&=
\arg\max_{\tilde{\sigma}\in\mathcal{S}}
\left|
\Delta\bar{D}_{\mathrm{tune}}(\tilde{\sigma})
\right|,
\label{eq:sigma_optim}
\end{align}
where $\mathcal{S}$ is the set of candidate noise levels. The sign of $\Delta\bar{D}_{\mathrm{tune}}(\tilde{\sigma}^{*})$ then determines the membership-score orientation: a positive gap indicates that higher divergence corresponds to membership, whereas a negative gap indicates that lower divergence corresponds to membership.

\section{Blind-Classifier Filtering}
\label{app:blind_classifier_filtering}

Motivated by evidence that differences between member and non-member data sources can confound membership inference evaluations~\cite{das2025blind,zhu2025revisiting}, we filter our self-constructed Endoscopy candidate pools using a classifier with no access to the target model. This procedure reduces the extent to which an attack can exploit image-source differences in place of target-model behavior.

We represent each image using the 512-dimensional global-average-pooled features preceding the classification head of a frozen ImageNet-pretrained ResNet-18~\cite{he2016deep}. We train a logistic regression classifier to distinguish candidate members from non-members using five-fold stratified cross-validation. We then obtain an out-of-fold predicted probability for each image and rank images within each class by the classifier's confidence in their true label. Images assigned lower confidence are considered harder for the blind classifier to distinguish.

We retain the hardest fraction of images from each class and adjust the keep-fraction until the classifier's AUC on the retained pools falls below $0.52$. The selected keep-fraction is $30\%$, reducing the pool from 18{,}623 candidate images to 5{,}588 retained images: 3{,}538 members and 2{,}050 non-members. The blind classifier's five-fold cross-validated AUC is $0.954\pm0.002$ on the full pools which reduces to $0.465\pm0.007$ on the retained pools. We report the latter as a post-filter diagnostic of separability under this blind baseline. Balanced target sets are subsequently sampled from the retained member and non-member pools.

\section{Additional Results}
\label{app:Additional Results}

\subsection{Raw Perturbed Images}
\begin{figure*}[!t]
    \centering
    \includegraphics[width=1\linewidth]{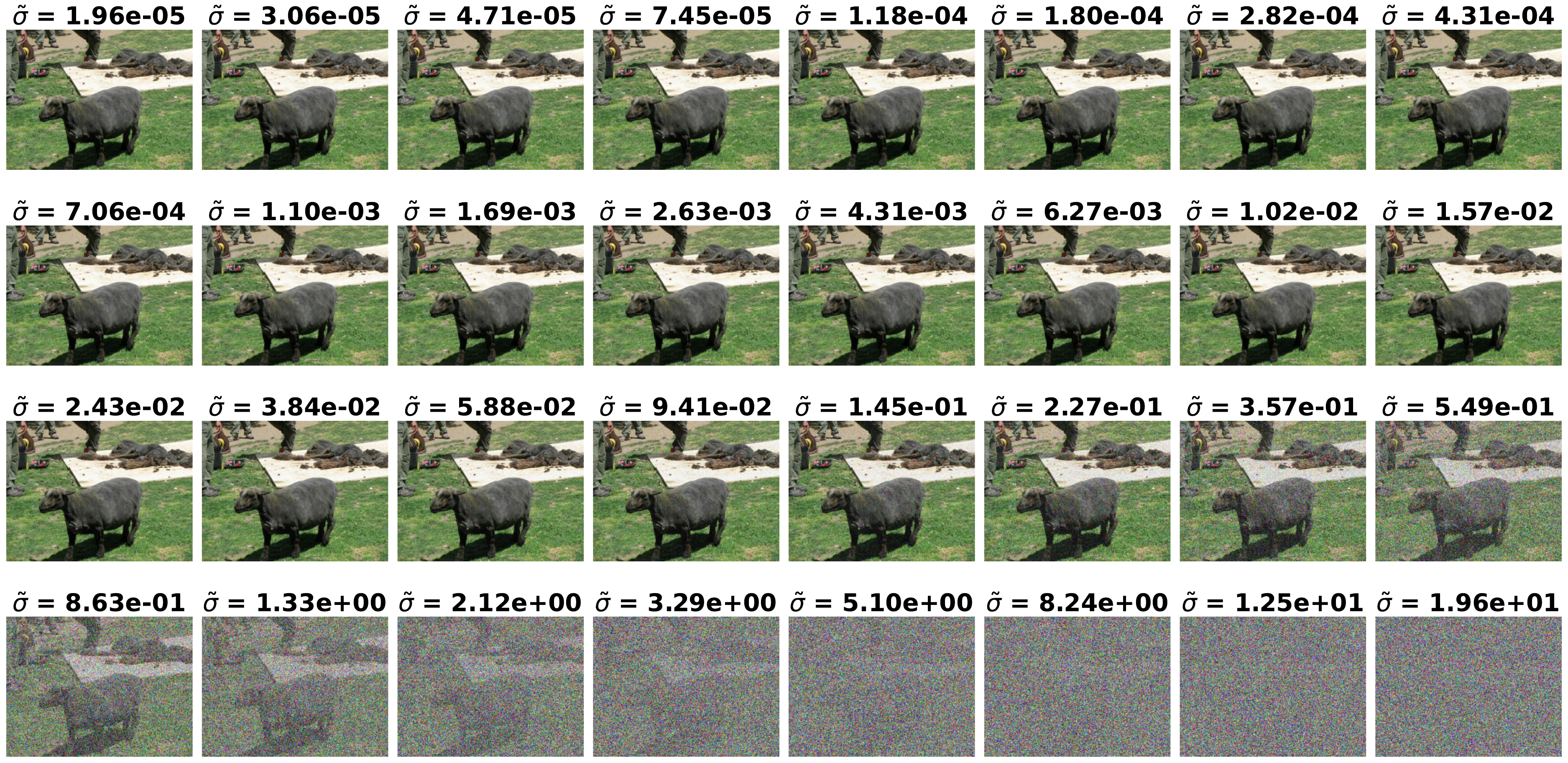}
    \caption{Raw Image Perturbations Used for Noise-level Tuning \autoref{sec:5_attack_methodology} and Divergence Analyses \autoref{sec:4_the_low_std_region}.}
    \label{fig:raw_images}
\end{figure*}

We show example perturbed versions of an input image used in our membership evaluation for the MIA (Figure \ref{fig:raw_images}); for each image, the corresponding normalized noise standard deviation $\tilde{\sigma}$ is indicated.

As discussed in section \ref{sec:5_attack_methodology}, we apply noise by independently sampling, for each pixel, from a Gaussian distribution with standard deviation $\tilde{\sigma}$, adding the resulting value to the pixel intensities (0–255), and then clipping. As shown in Figure \ref{fig:raw_images}, this procedure produces a noisy image in which an increasing fraction of pixels saturate to complete noise due to clipping, while the remaining pixels become progressively more faded. Notably, in looking at the regime II range of $\tilde{\sigma}$ values which often proved most optimal for our target sets in noise-level tuning (1.10e-03 to 2.43e-02), we qualitatively observe that these region adds noise that is indistinguishable to the human eye, demonstrating the minute sensitivity of VLLMs.

\subsection{Raw Divergence Plots}
\begin{figure*}[h]
    \centering
    \begin{subfigure}{0.32\textwidth}
        \centering
        \includegraphics[width=\linewidth]{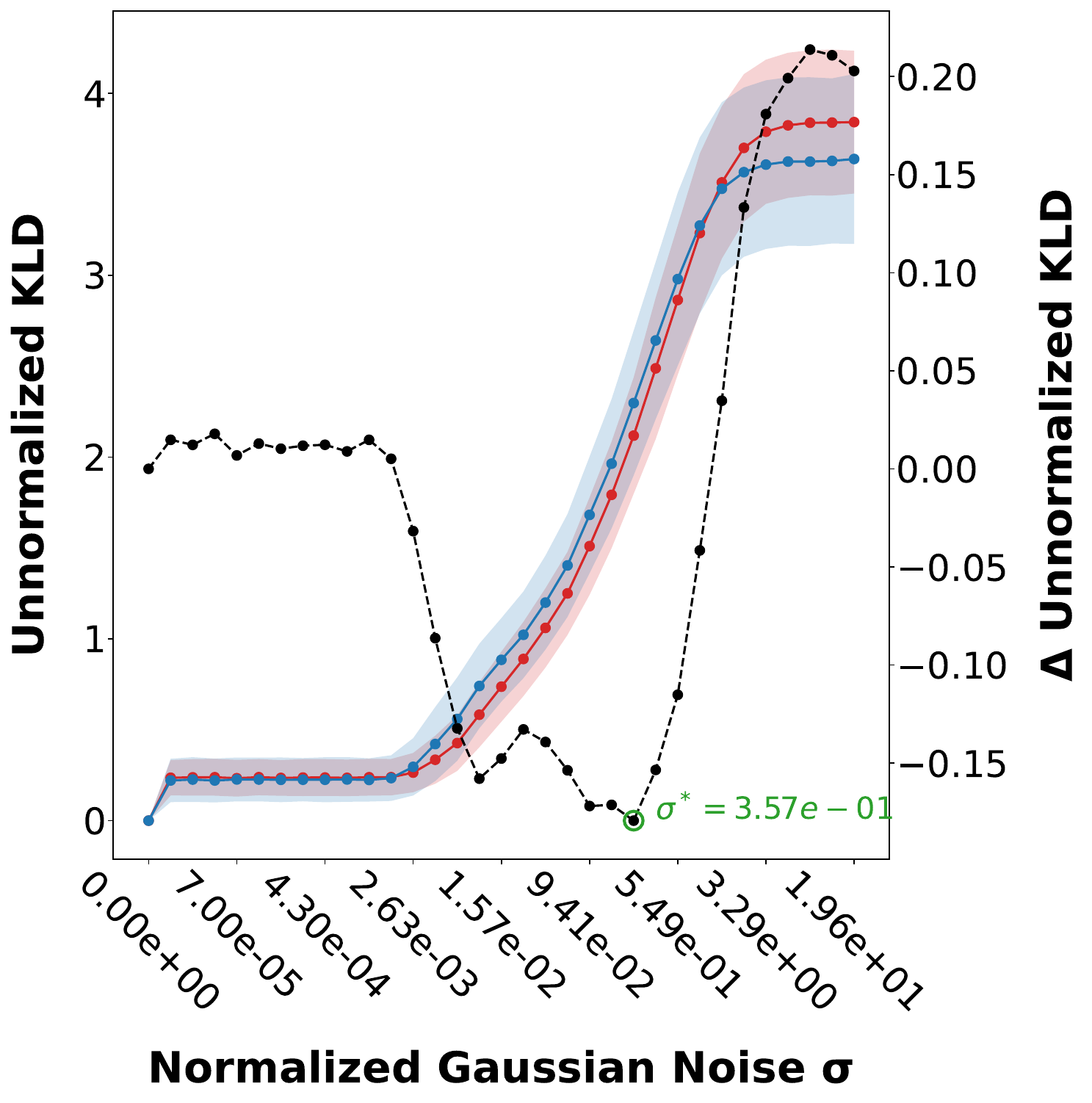}
        \caption{Unnoramlised KL-Divergence}
    \end{subfigure}\hfill
    \begin{subfigure}{0.32\textwidth}
        \centering
        \includegraphics[width=\linewidth]{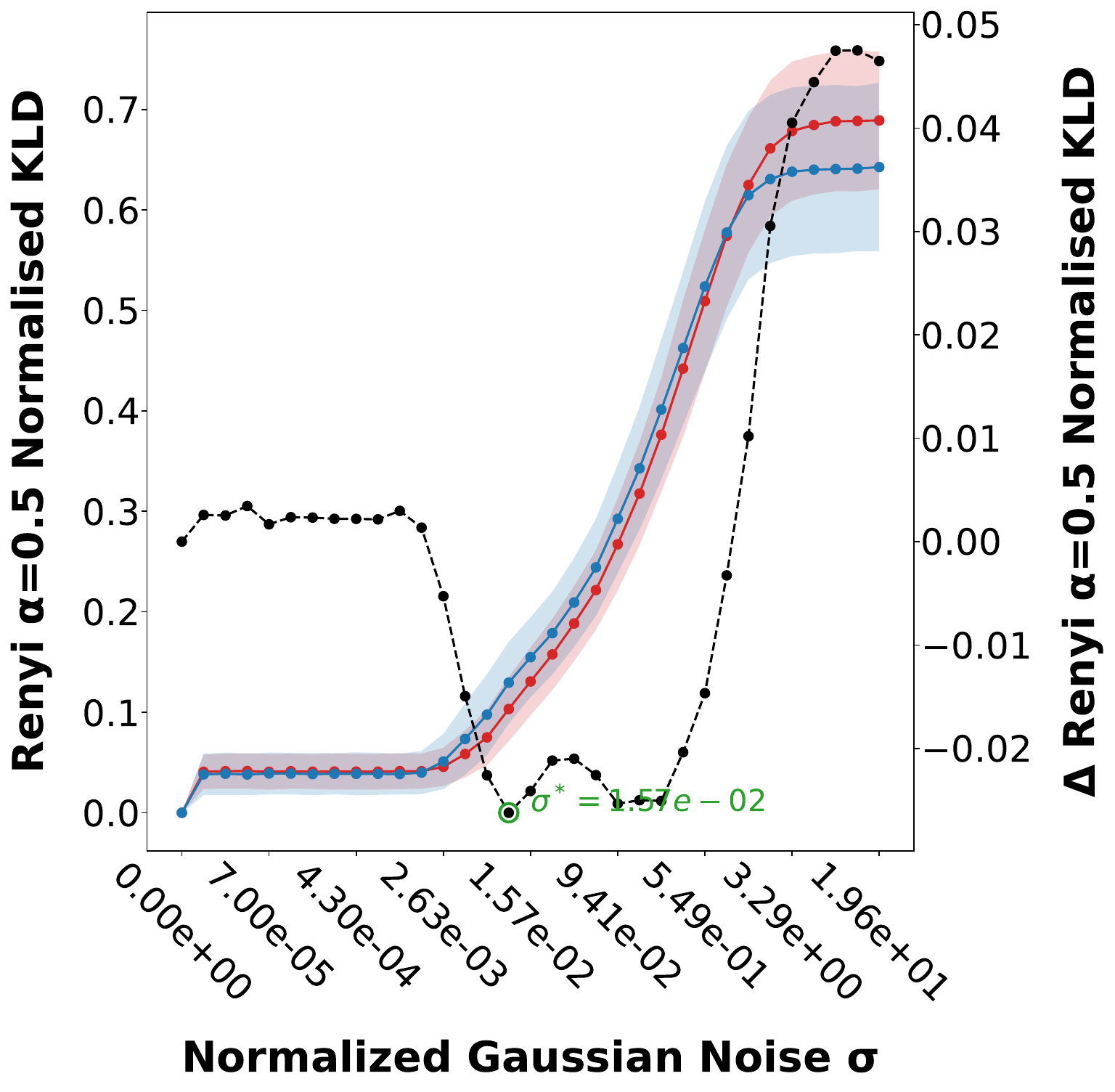}
        \caption{Renyi $\alpha=0.5$ Normalised KL-Divergence}
    \end{subfigure}\hfill
    \begin{subfigure}{0.32\textwidth}
        \centering
        \includegraphics[width=\linewidth]{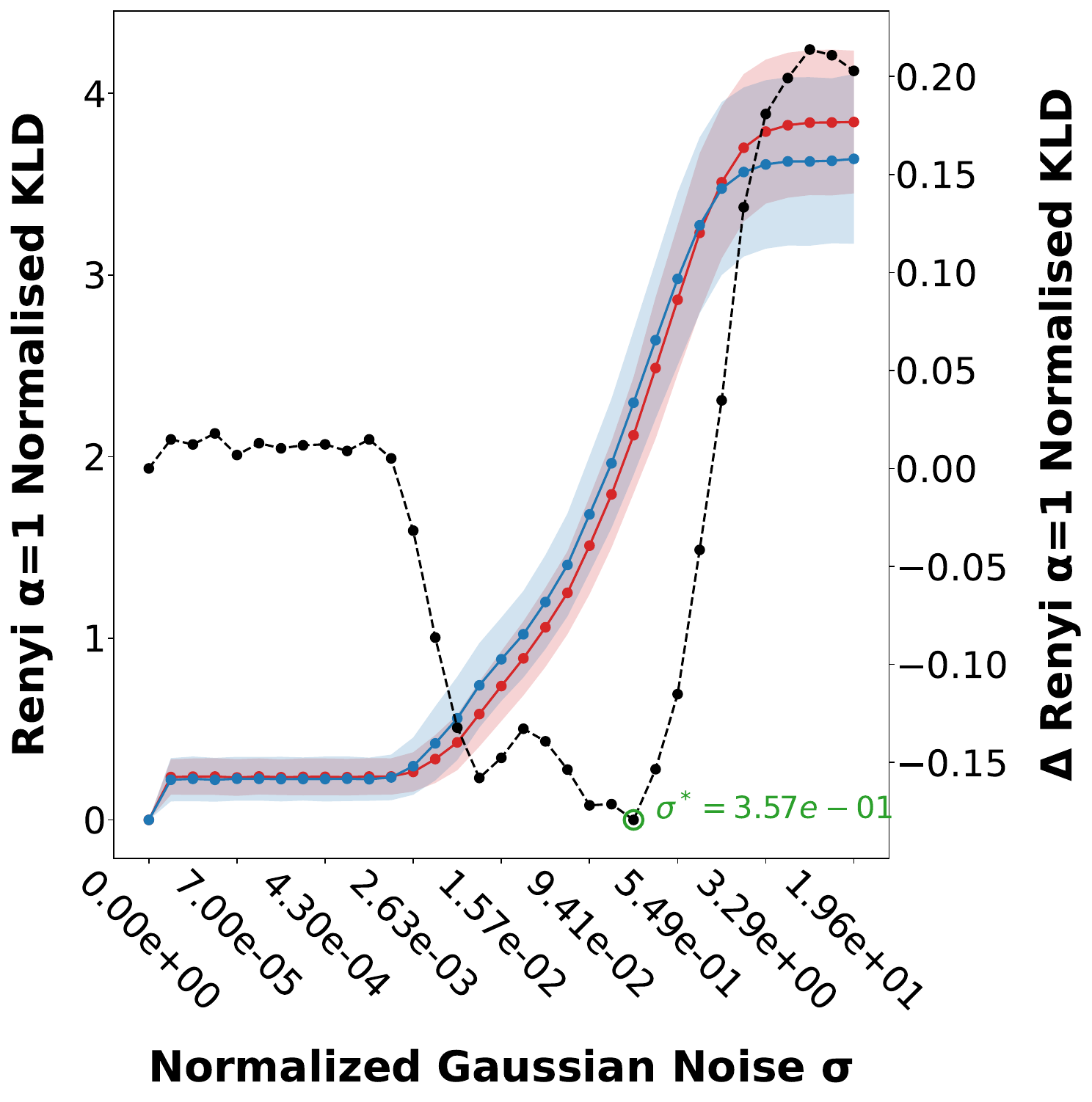}
        \caption{Renyi $\alpha=1$ Normalised KL-Divergence}
    \end{subfigure}\hfill

    \vspace{0.8em}
    
    \begin{subfigure}{0.32\textwidth}
        \centering
        \includegraphics[width=\linewidth]{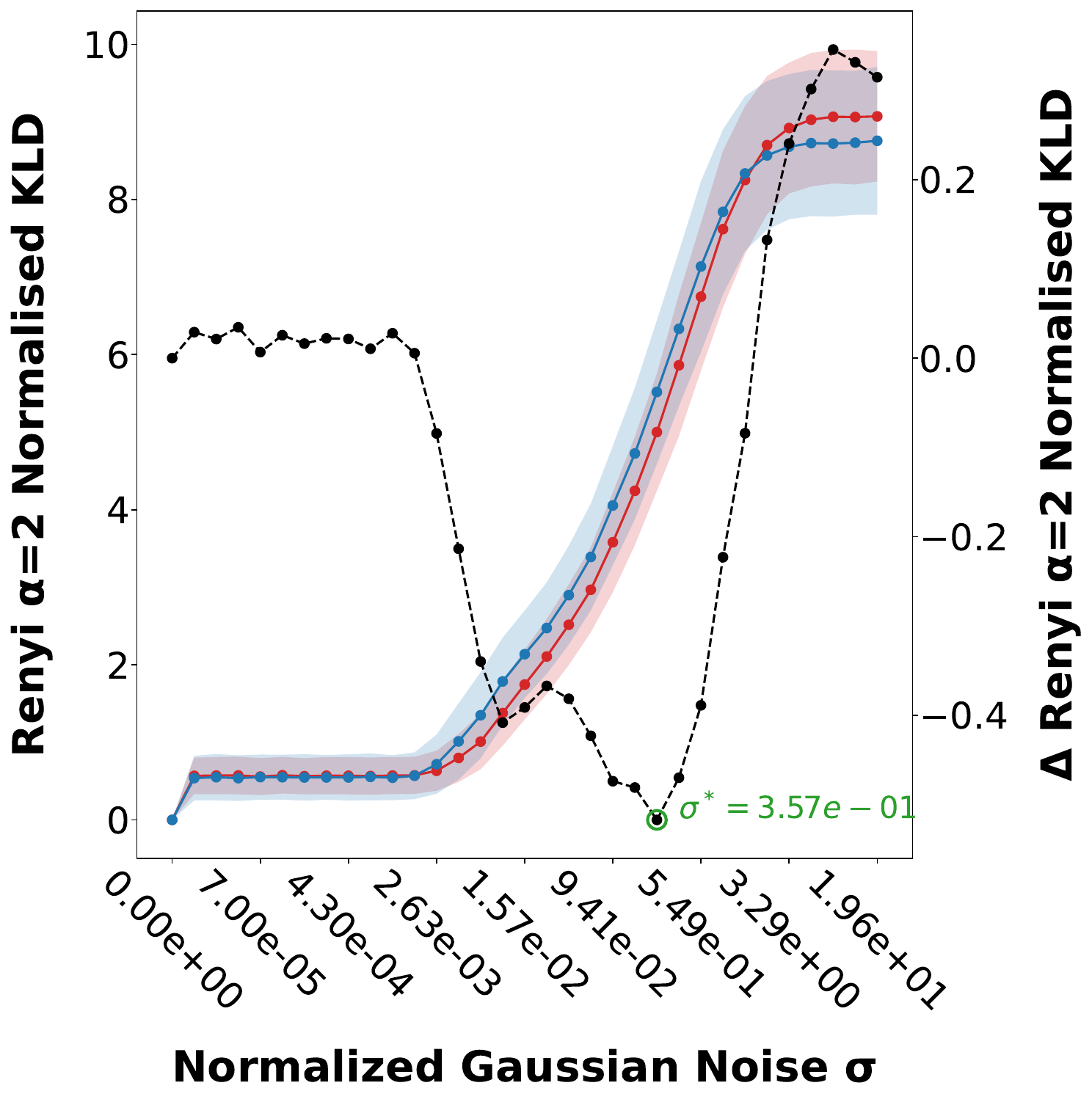}
        \caption{Renyi $\alpha=2$ Normalised KL-Divergence}
    \end{subfigure}\hfill
    \begin{subfigure}{0.32\textwidth}
        \centering
        \includegraphics[width=\linewidth]{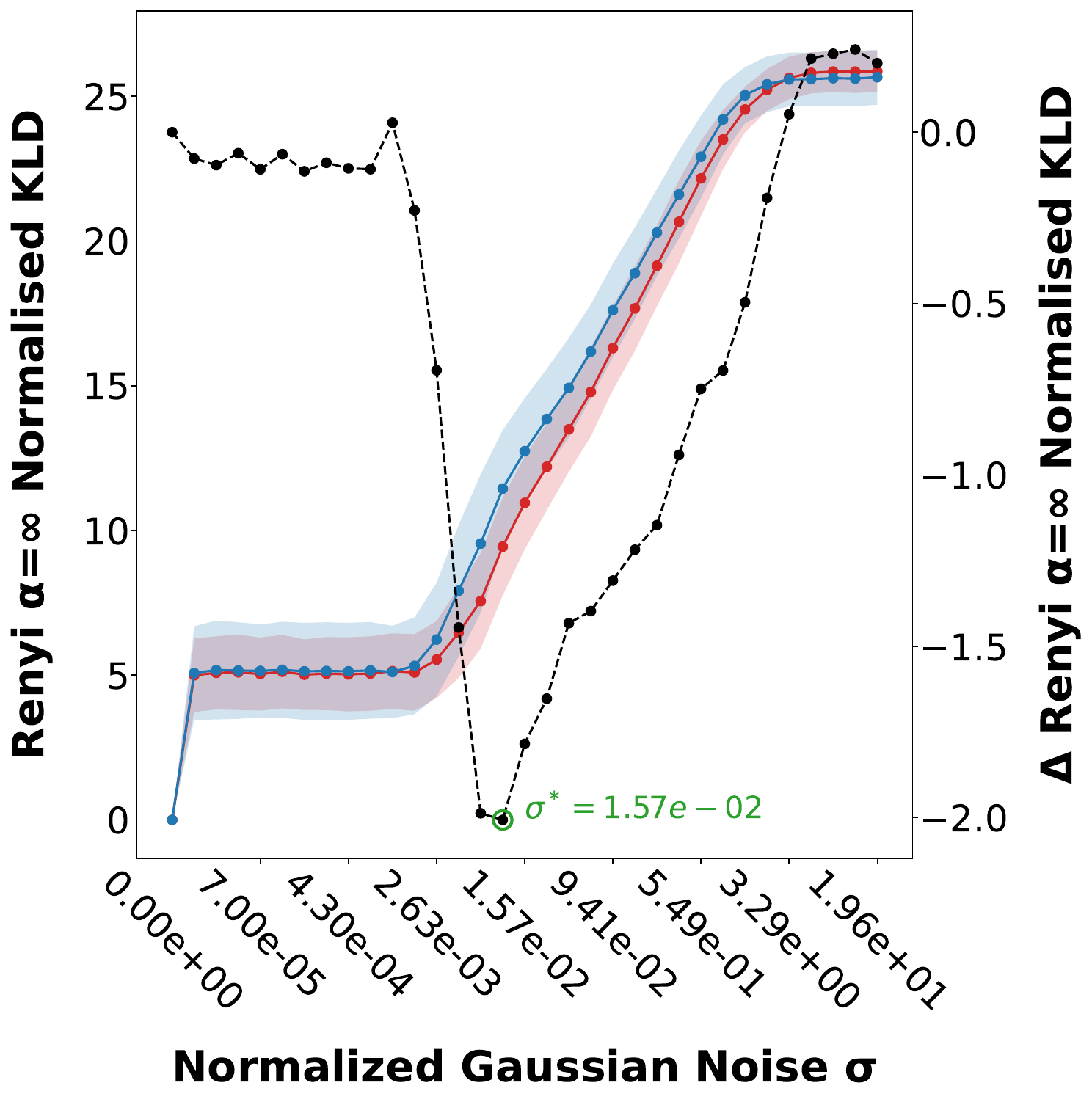}
        \caption{Renyi $\alpha=\infty$ Normalised KL-Divergence}
    \end{subfigure}\hfill
    \begin{subfigure}{0.32\textwidth}
        \centering
        \includegraphics[width=\linewidth]{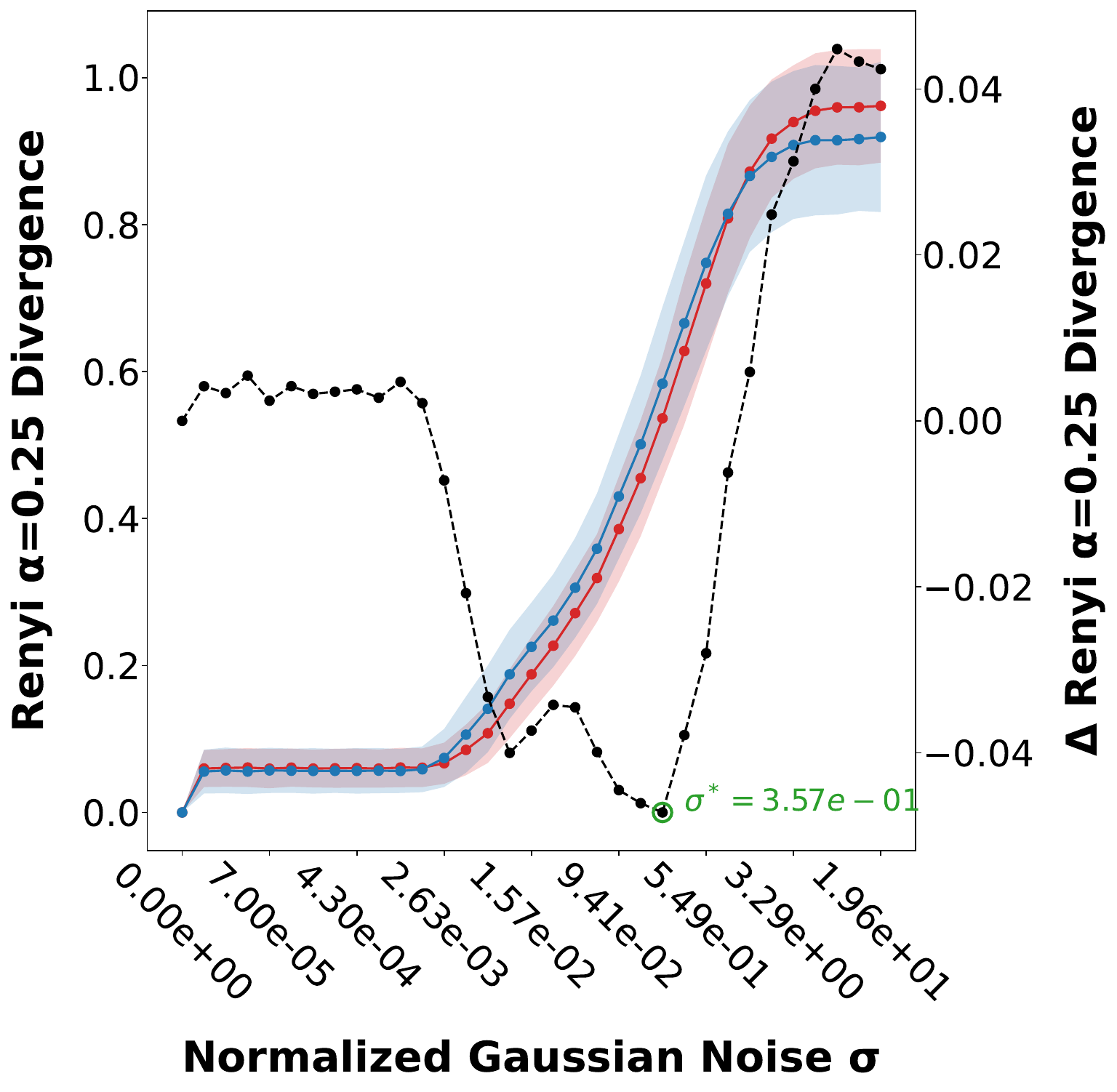}
        \caption{Renyi $\alpha=0.25$ Divergence}
    \end{subfigure}

    \vspace{0.5em}

    \begin{subfigure}{0.32\textwidth}
        \centering
        \includegraphics[width=\linewidth]{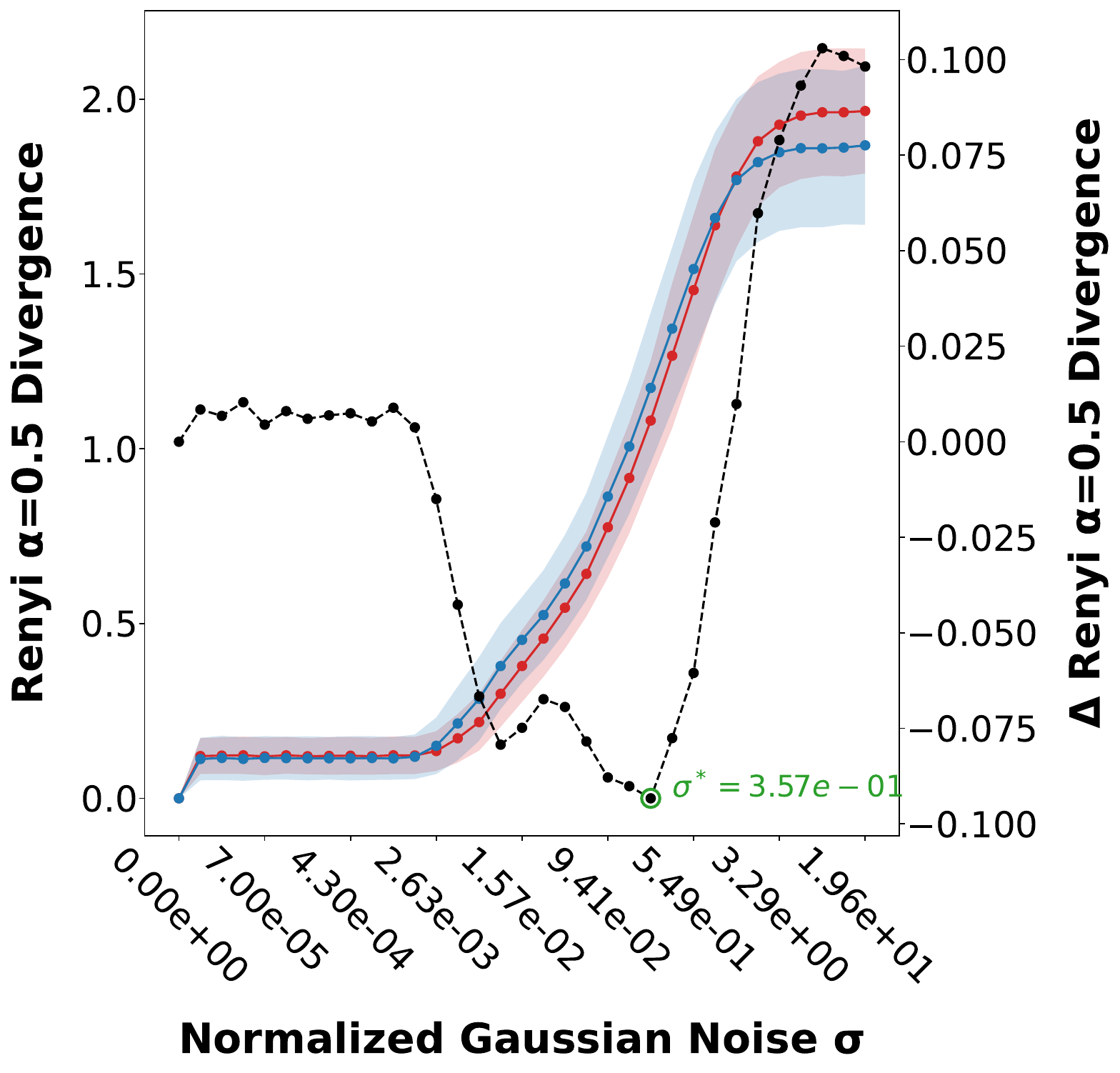}
        \caption{Renyi $\alpha=0.5$ Divergence}
    \end{subfigure}\hfill
    \begin{subfigure}{0.32\textwidth}
        \centering
        \includegraphics[width=\linewidth]{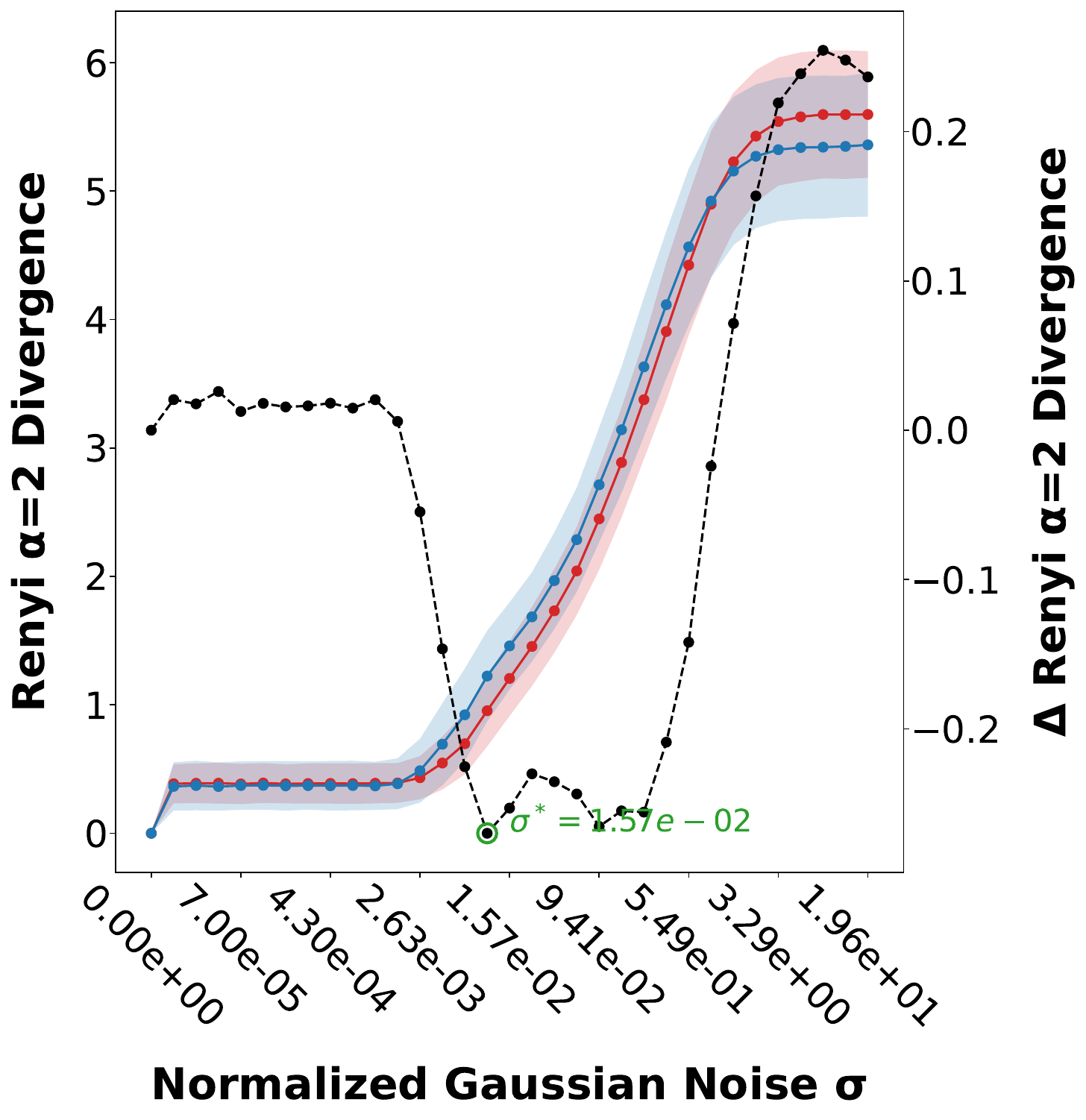}
        \caption{Renyi $\alpha=2$ Divergence}
    \end{subfigure}\hfill
    \begin{subfigure}{0.32\textwidth}
        \centering
        \includegraphics[width=\linewidth]{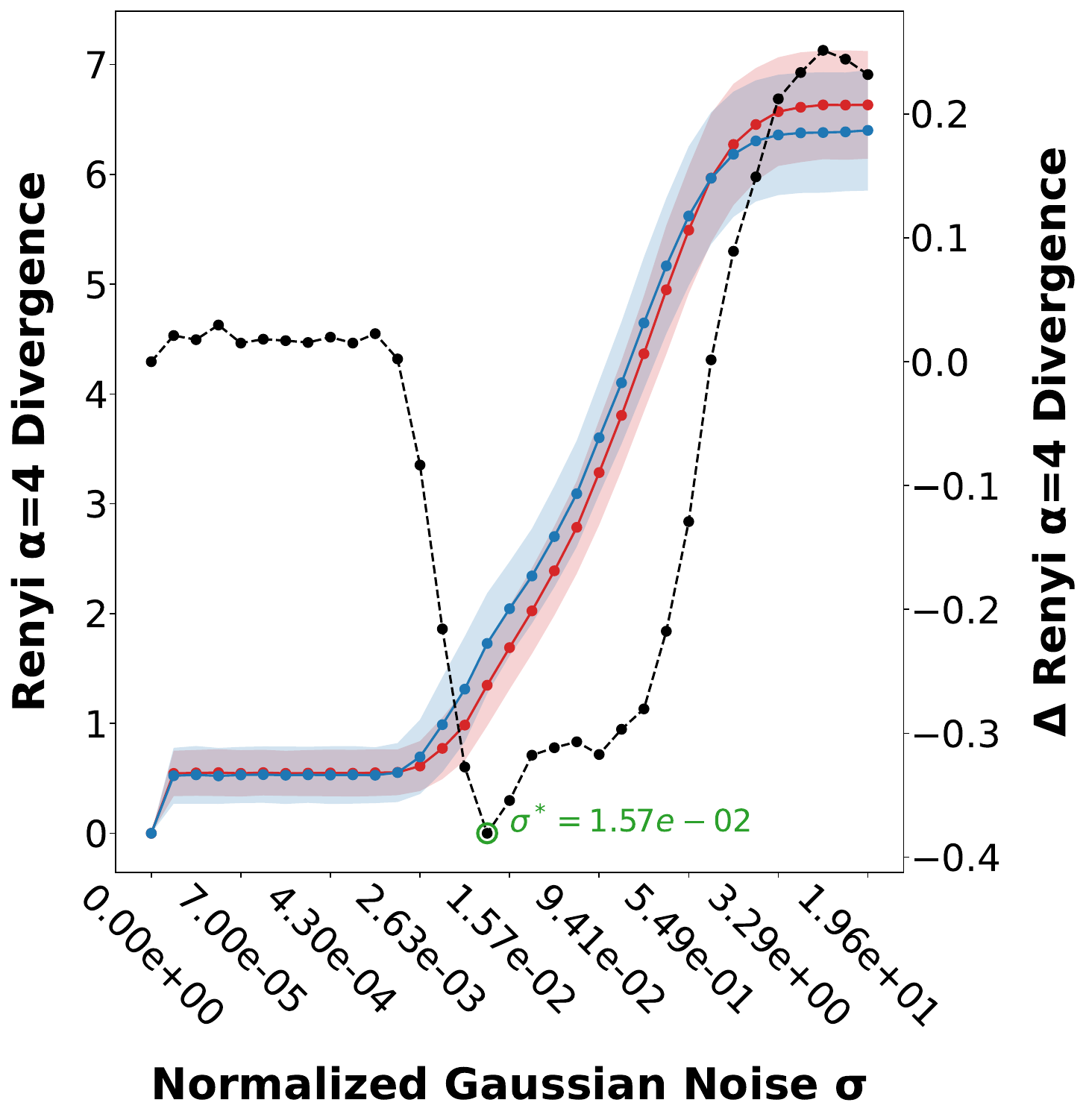}
        \caption{Renyi $\alpha=4$ Divergence}
    \end{subfigure}

    \caption{Raw divergence curves for all tested normalization for a flickr target subset members (red), non-members (blue), and member-non-member difference (black)  against the LLaVA model.}
    \label{fig:raw_divs_llava}
\end{figure*}

\begin{figure*}[h]
    \centering

    \begin{subfigure}{0.32\textwidth}
        \centering
        \includegraphics[width=\linewidth]{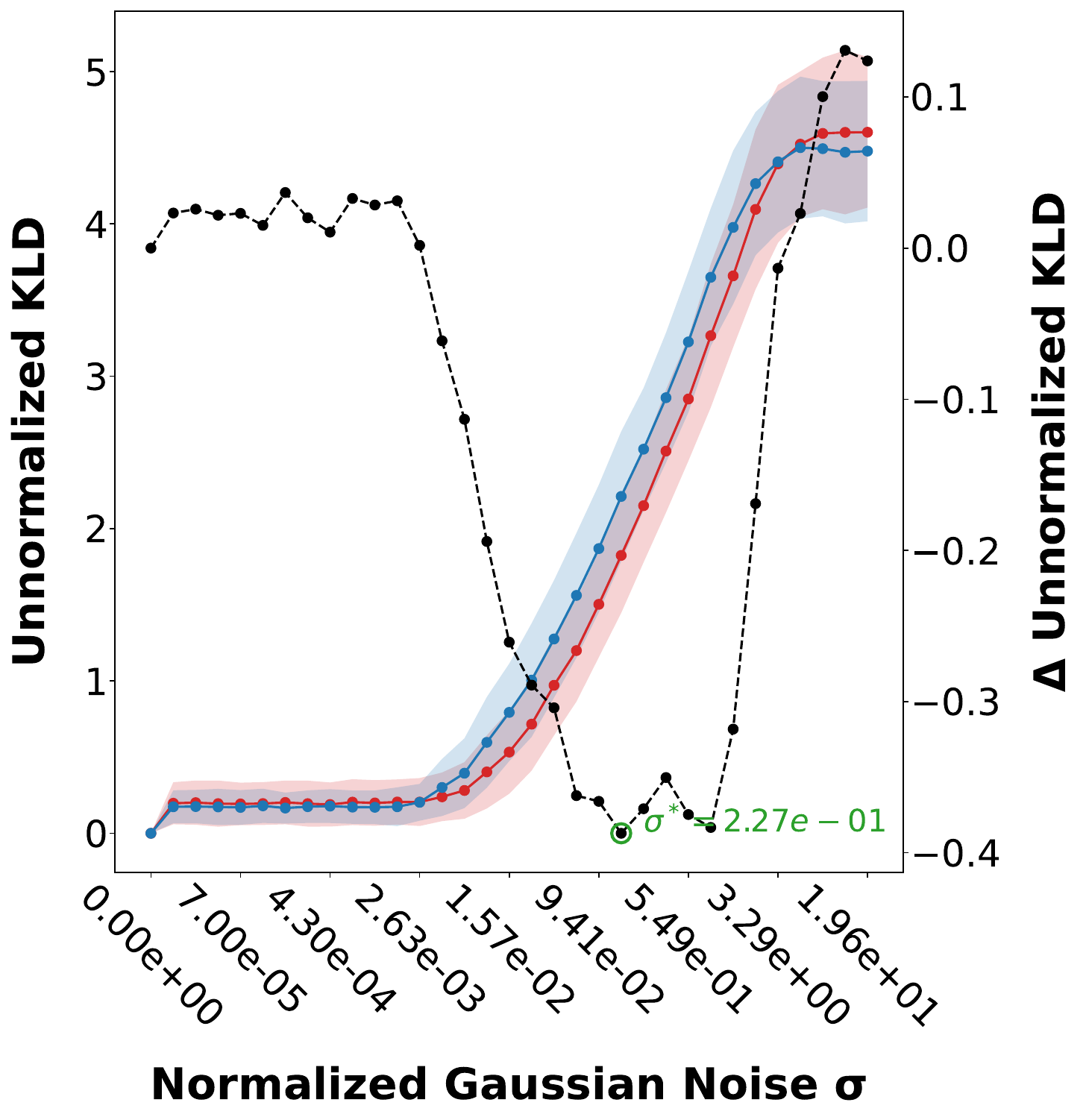}
        \caption{Unnoramlised KL-Divergence}
    \end{subfigure}\hfill
    \begin{subfigure}{0.32\textwidth}
        \centering
        \includegraphics[width=\linewidth]{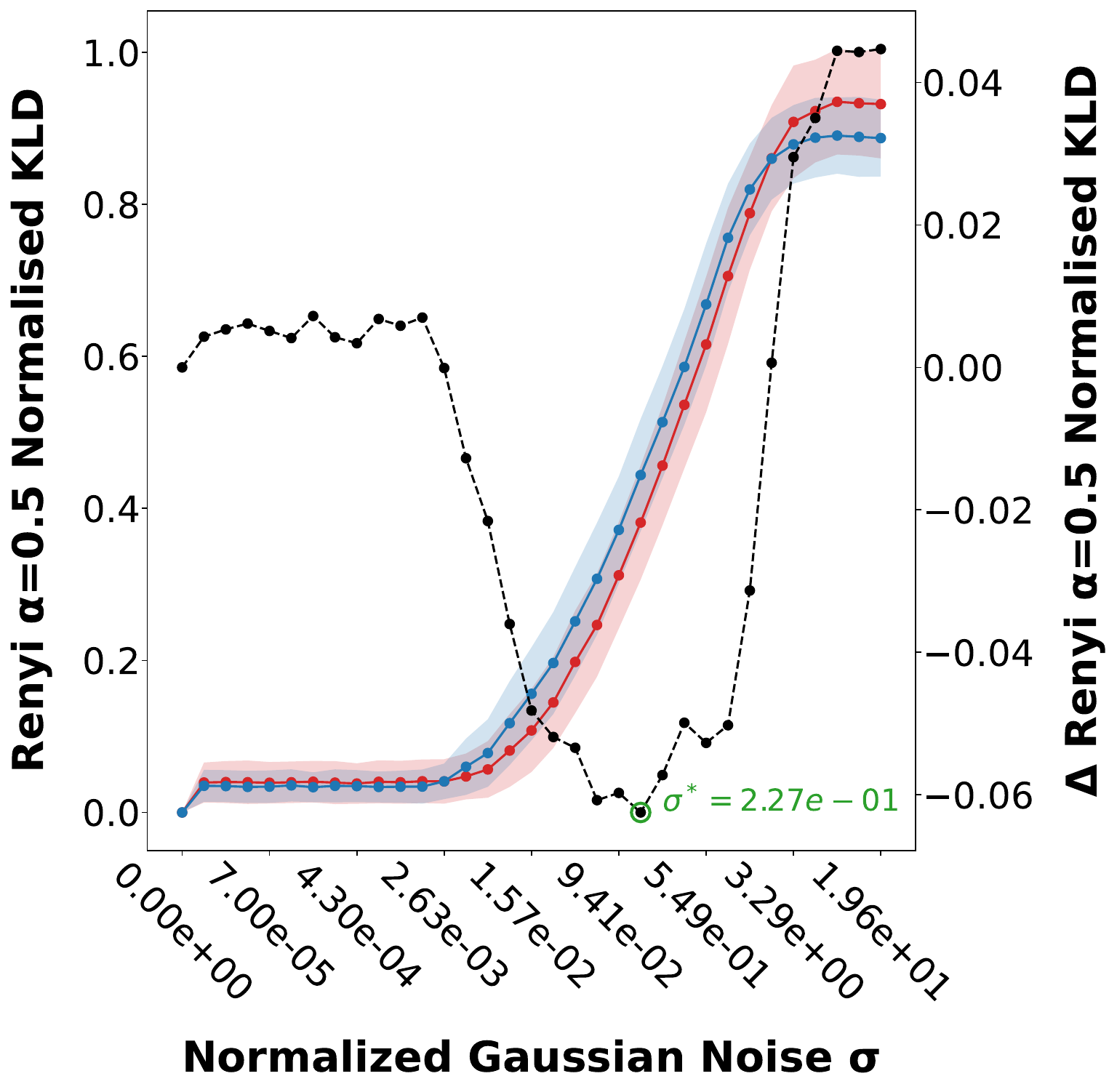}
        \caption{Renyi $\alpha=0.5$ Normalised KL-Divergence}
    \end{subfigure}\hfill
    \begin{subfigure}{0.32\textwidth}
        \centering
        \includegraphics[width=\linewidth]{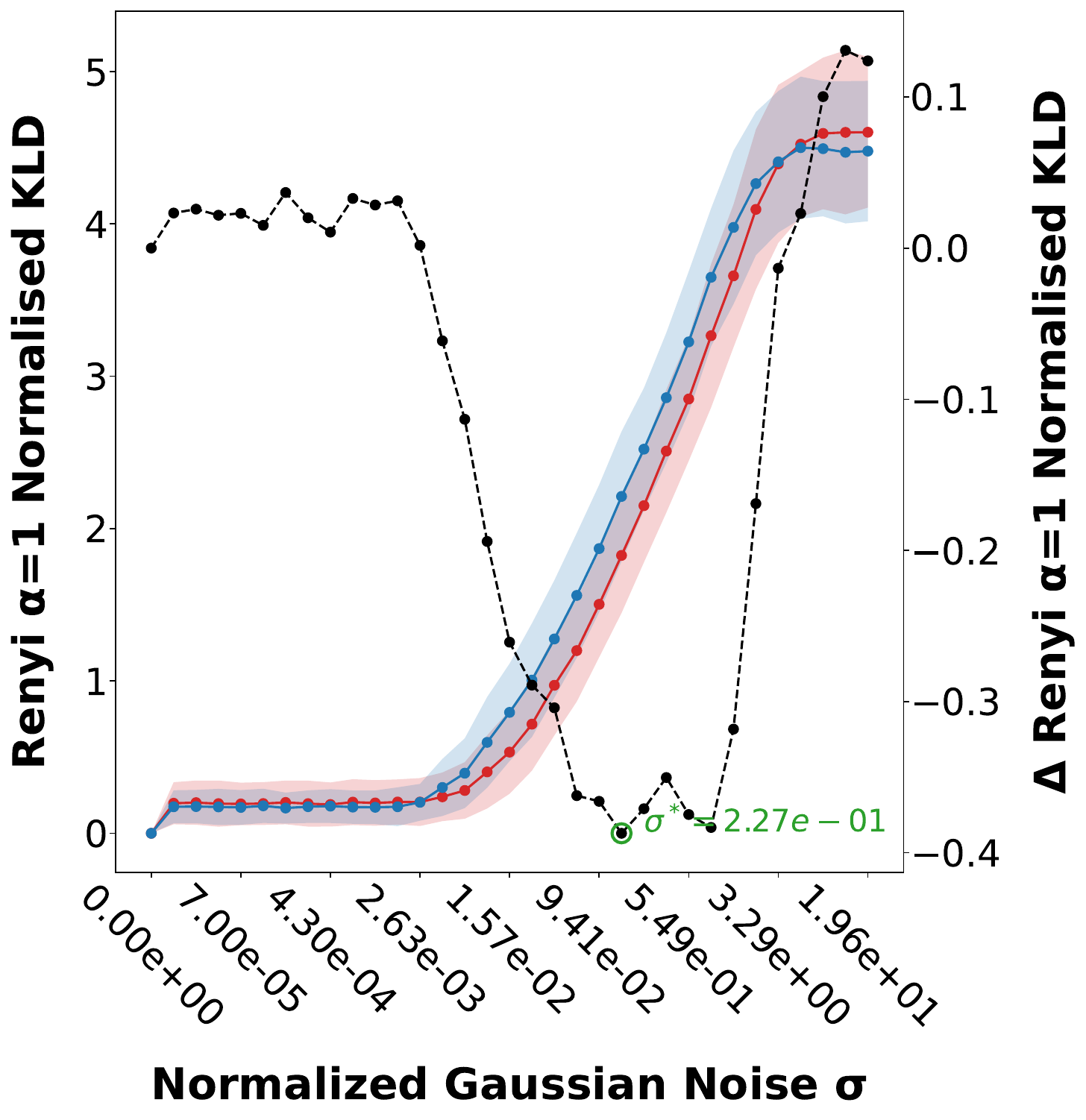}
        \caption{Renyi $\alpha=1$ Normalised KL-Divergence}
    \end{subfigure}\hfill

    \vspace{0.8em}
    
    \begin{subfigure}{0.32\textwidth}
        \centering
        \includegraphics[width=\linewidth]{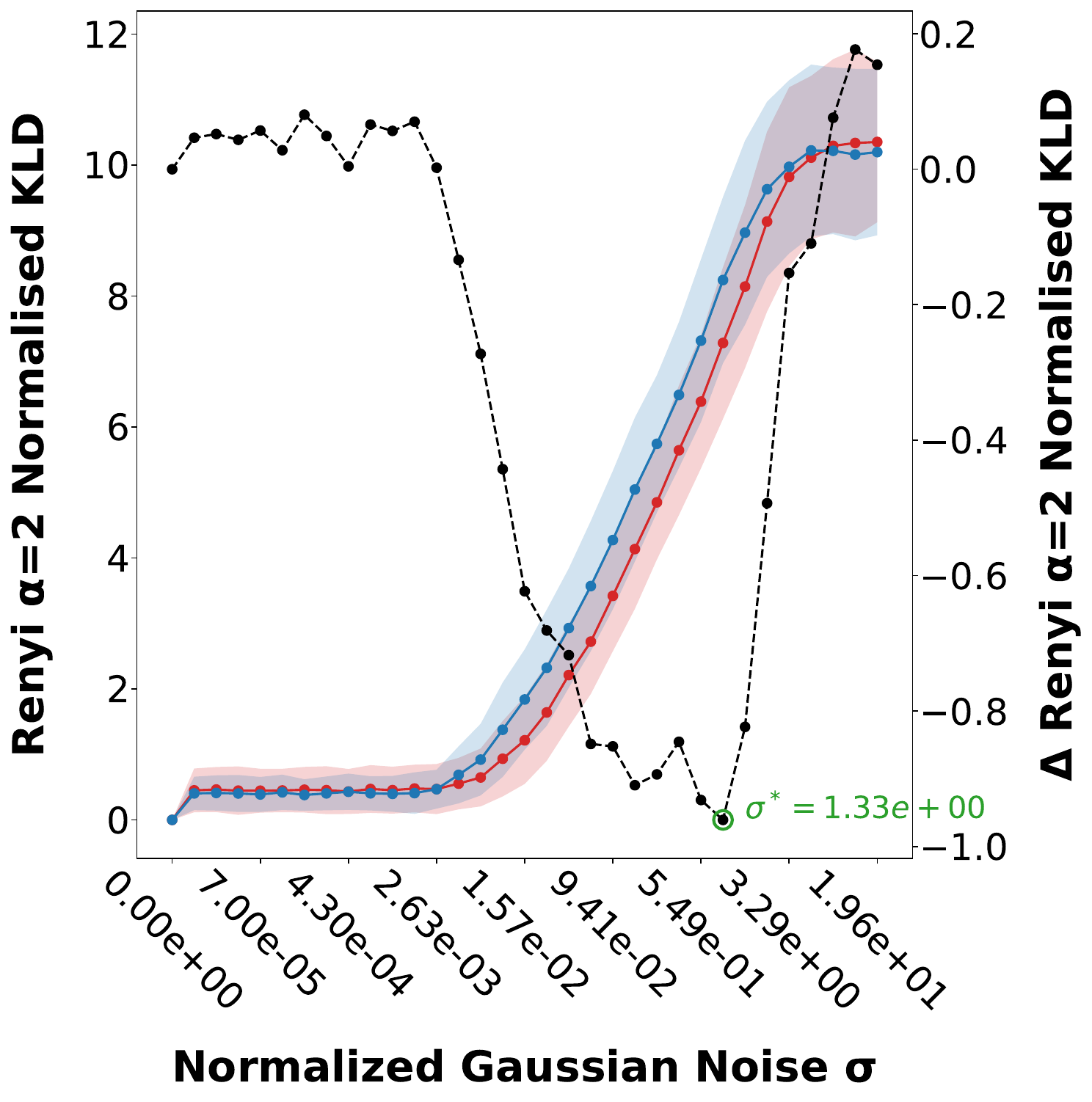}
        \caption{Renyi $\alpha=2$ Normalised KL-Divergence}
    \end{subfigure}\hfill
    \begin{subfigure}{0.32\textwidth}
        \centering
        \includegraphics[width=\linewidth]{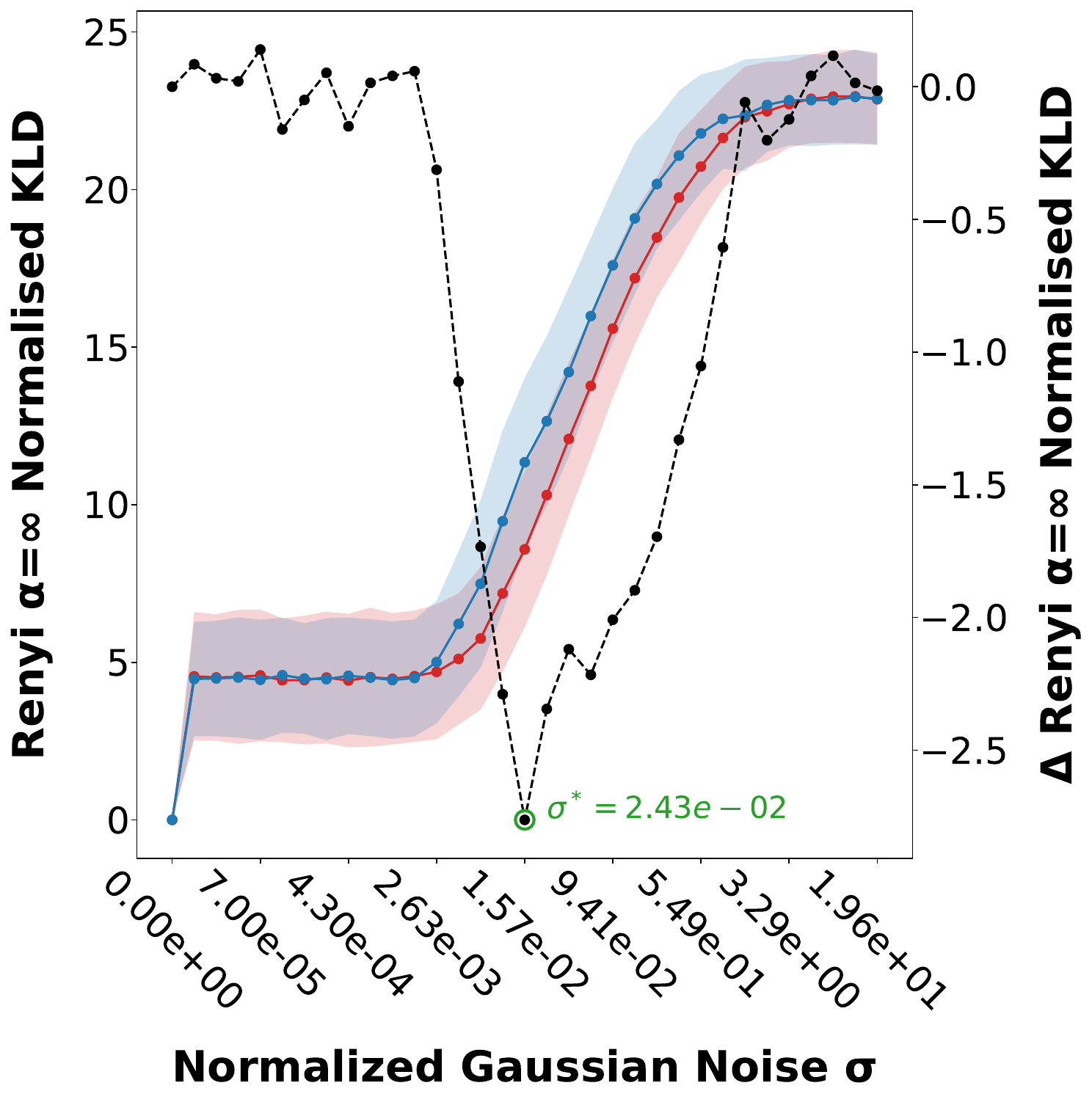}
        \caption{Renyi $\alpha=\infty$ Normalised KL-Divergence}
    \end{subfigure}\hfill
    \begin{subfigure}{0.32\textwidth}
        \centering
        \includegraphics[width=\linewidth]{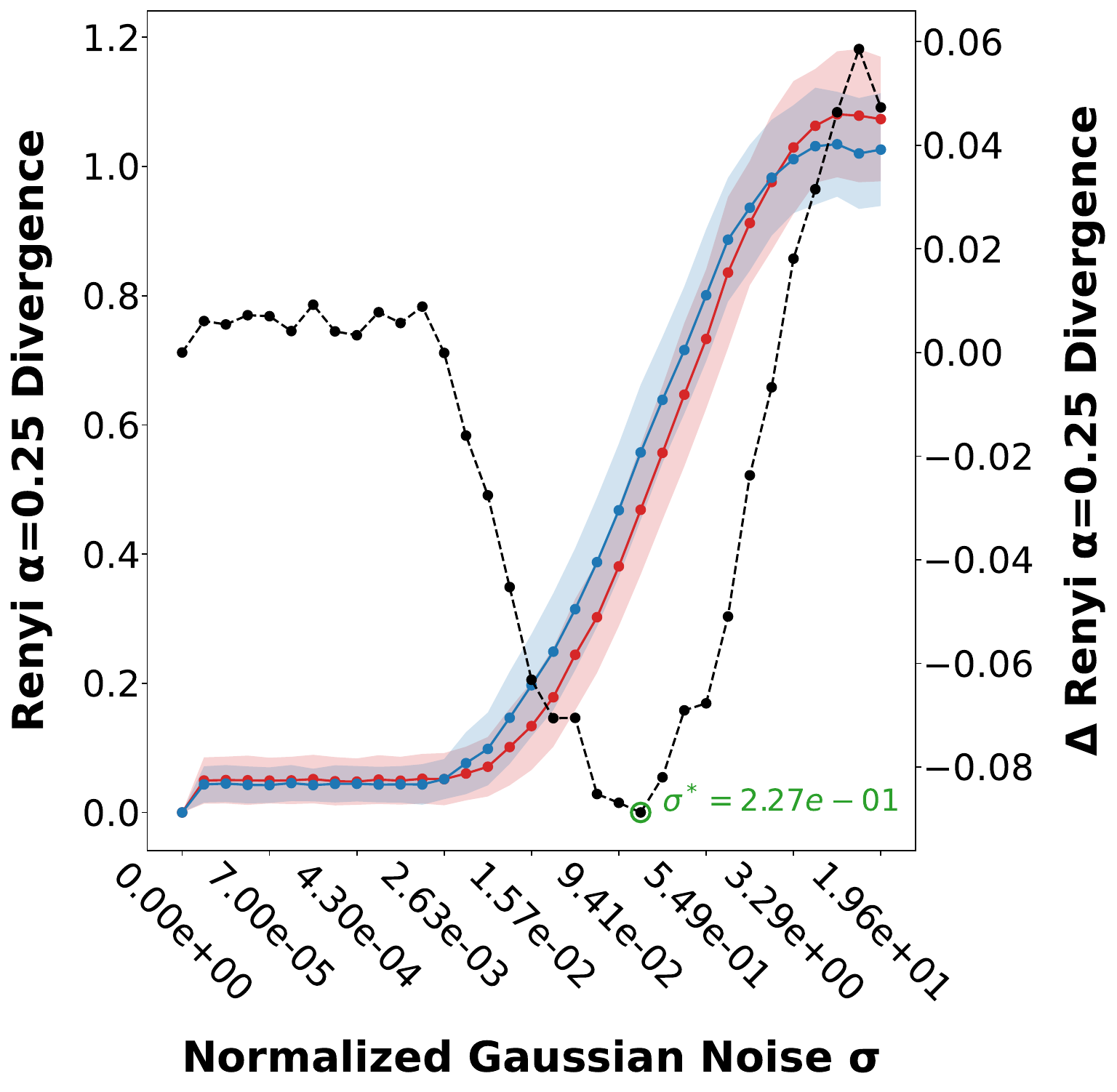}
        \caption{Renyi $\alpha=0.25$ Divergence}
    \end{subfigure}

    \vspace{0.5em}

    \begin{subfigure}{0.32\textwidth}
        \centering
        \includegraphics[width=\linewidth]{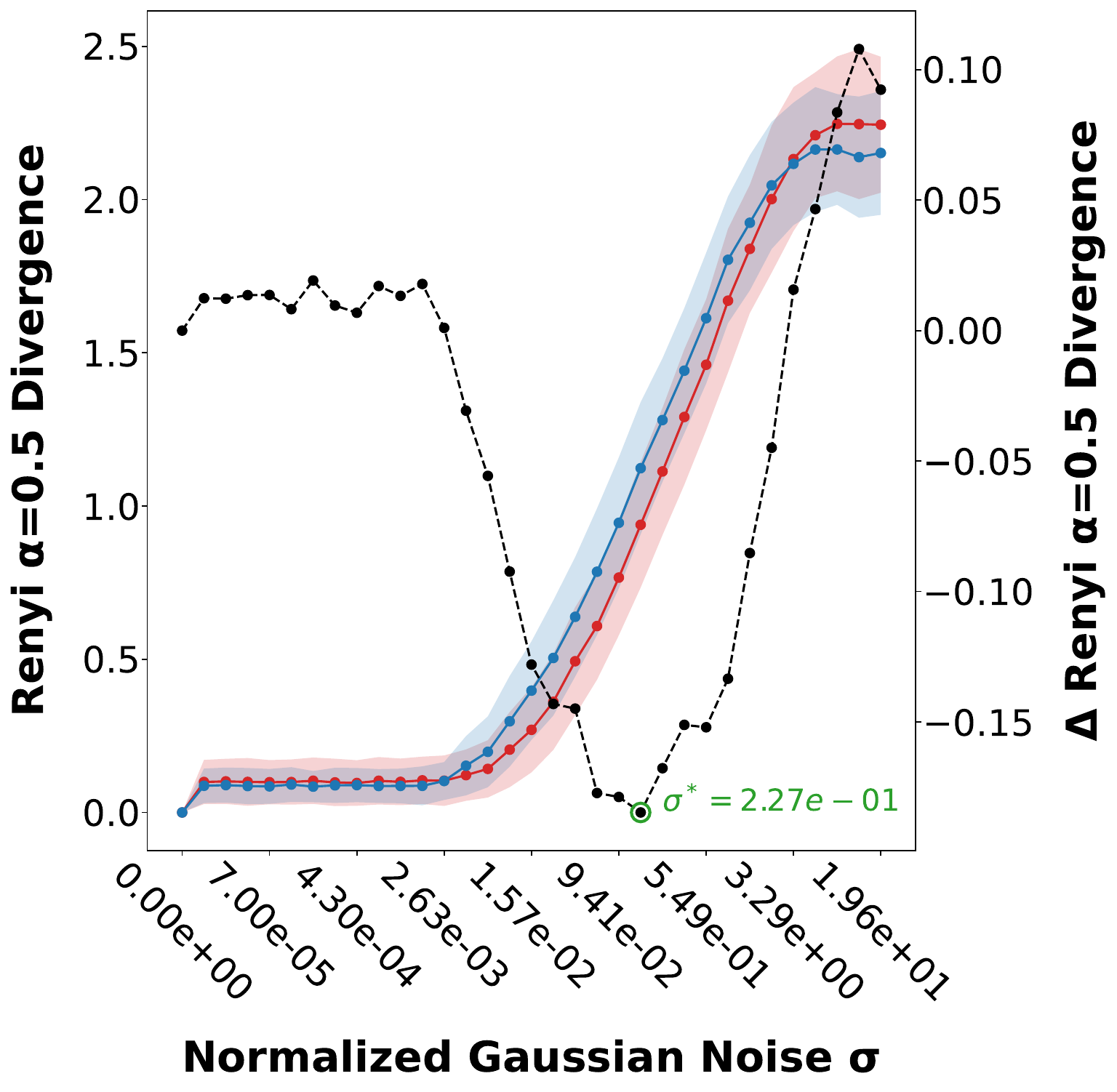}
        \caption{Renyi $\alpha=0.5$ Divergence}
    \end{subfigure}\hfill
    \begin{subfigure}{0.32\textwidth}
        \centering
        \includegraphics[width=\linewidth]{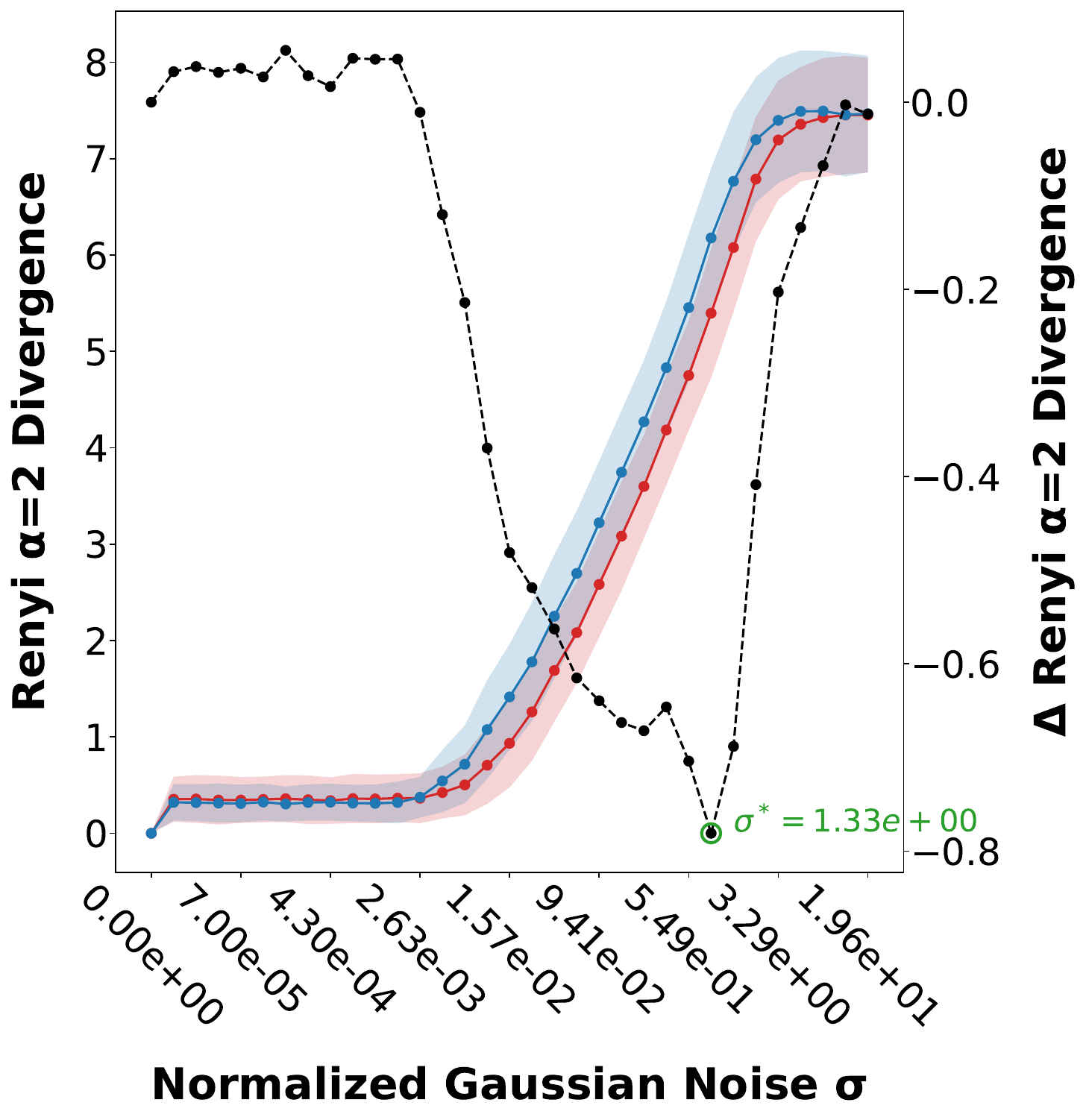}
        \caption{Renyi $\alpha=2$ Divergence}
    \end{subfigure}\hfill
    \begin{subfigure}{0.32\textwidth}
        \centering
        \includegraphics[width=\linewidth]{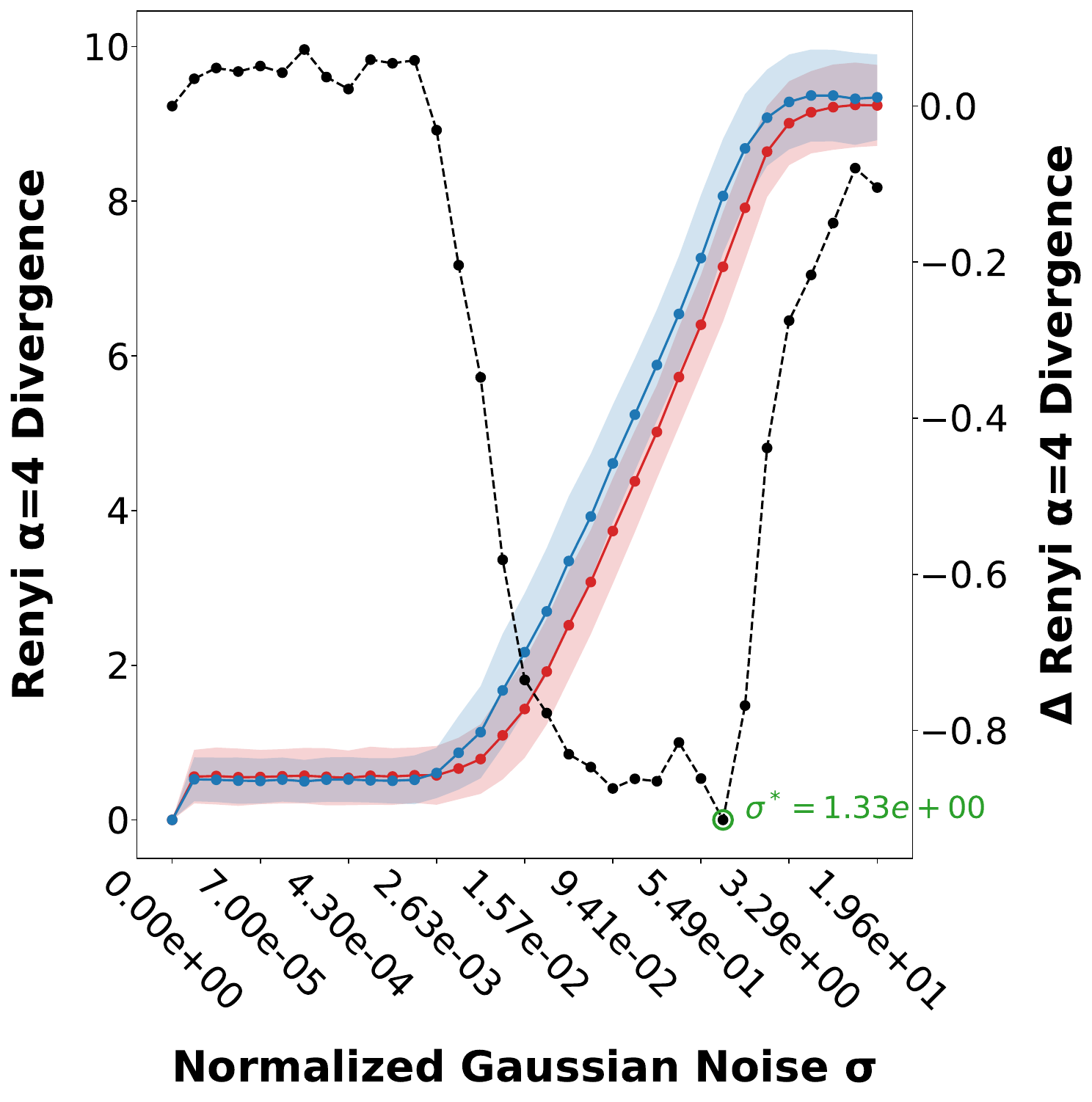}
        \caption{Renyi $\alpha=4$ Divergence}
    \end{subfigure}

    \caption{Raw divergence curves for all tested normalization for a flickr target subset members (red), non-members (blue), and member-non-member difference (black)  against the MiniGPT-4o model.}
    \label{fig:raw_divs_minigpt}
\end{figure*}

\begin{figure*}[h]
    \centering

    \begin{subfigure}{0.32\textwidth}
        \centering
        \includegraphics[width=\linewidth]{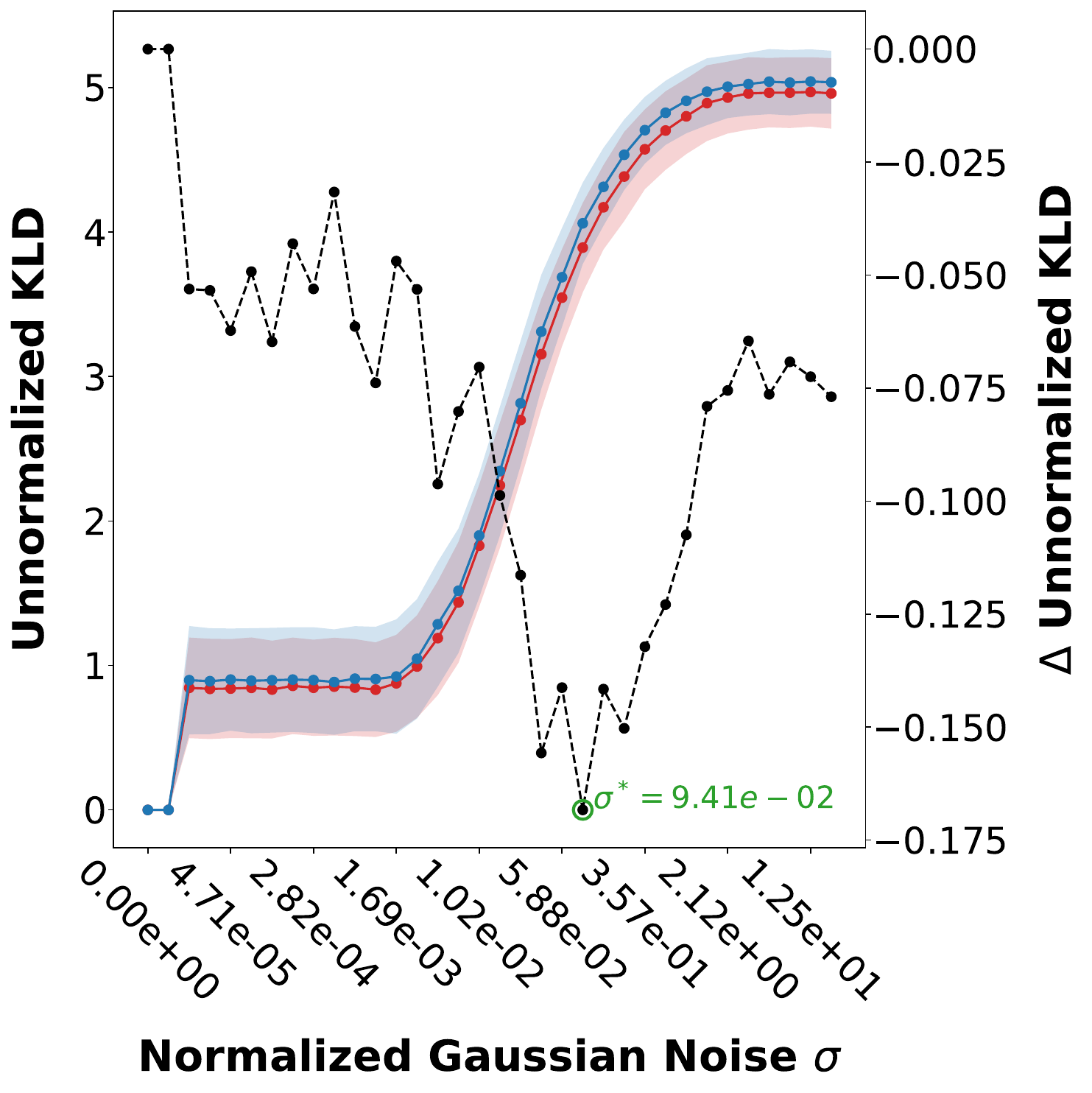}
        \caption{Unnoramlised KL-Divergence}
    \end{subfigure}\hfill
    \begin{subfigure}{0.32\textwidth}
        \centering
        \includegraphics[width=\linewidth]{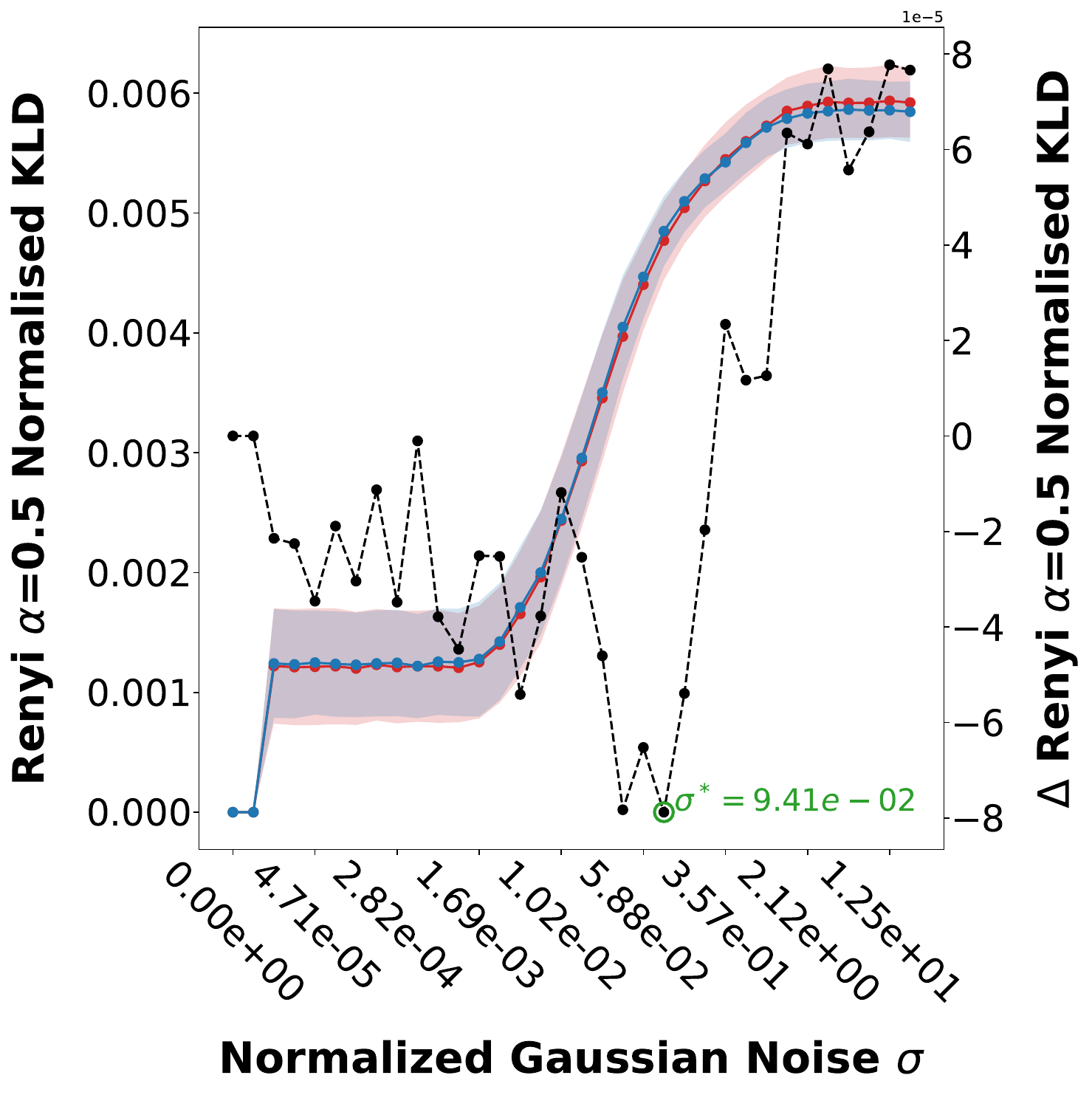}
        \caption{Renyi $\alpha=0.5$ Normalised KL-Divergence}
    \end{subfigure}\hfill
    \begin{subfigure}{0.32\textwidth}
        \centering
        \includegraphics[width=\linewidth]{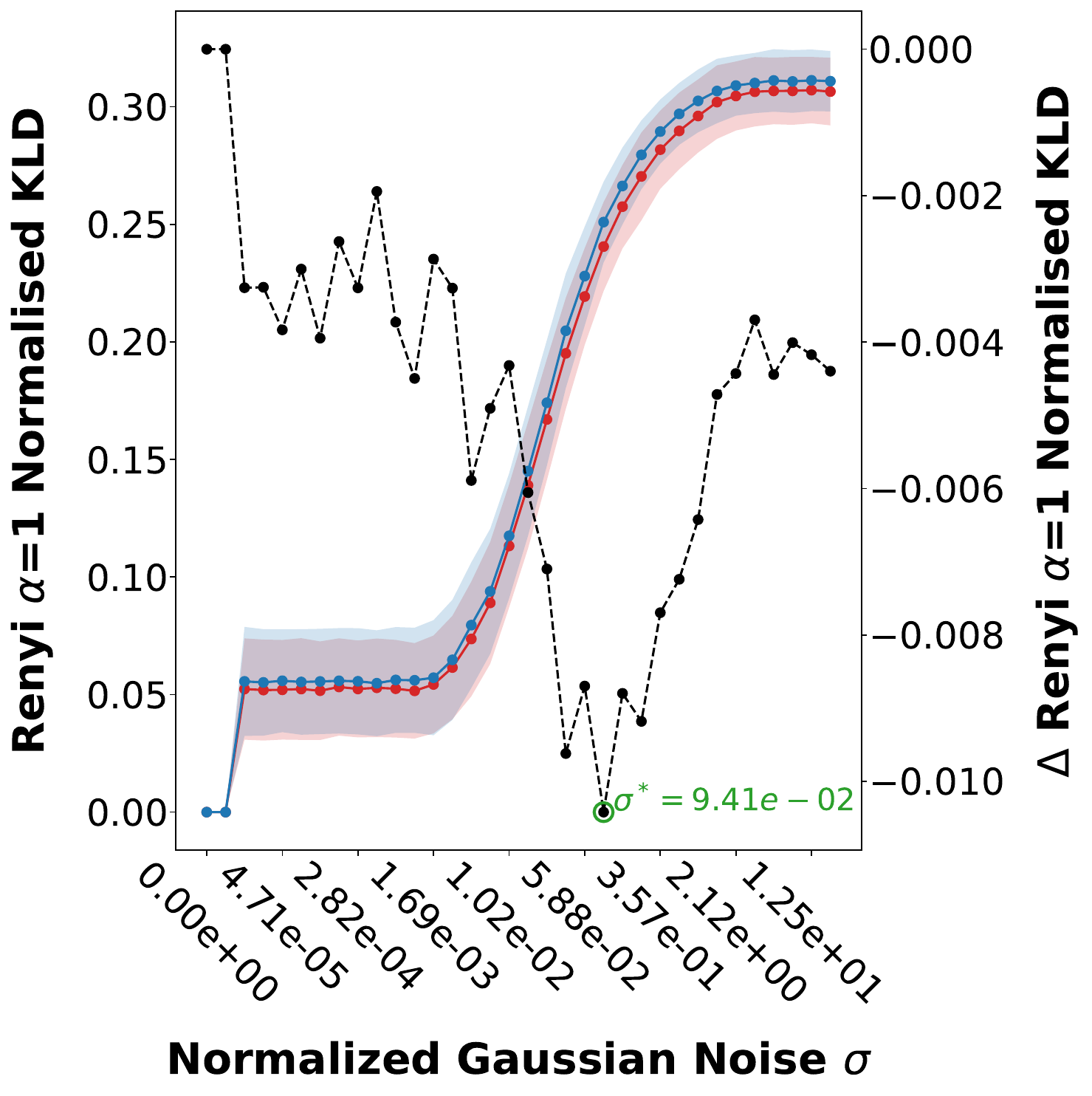}
        \caption{Renyi $\alpha=1$ Normalised KL-Divergence}
    \end{subfigure}\hfill

    \vspace{0.8em}
    
    \begin{subfigure}{0.32\textwidth}
        \centering
        \includegraphics[width=\linewidth]{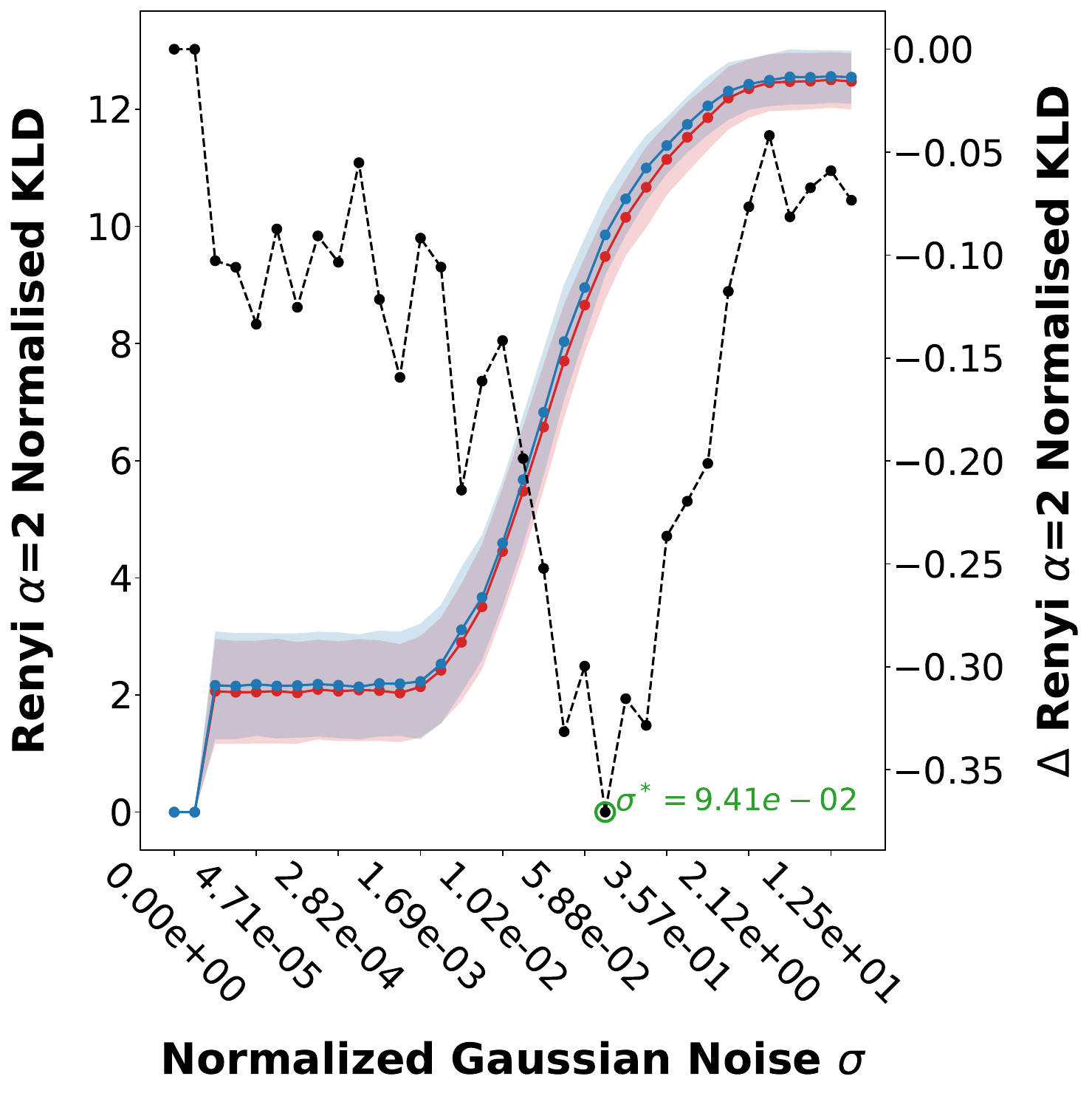}
        \caption{Renyi $\alpha=2$ Normalised KL-Divergence}
    \end{subfigure}\hfill
    \begin{subfigure}{0.32\textwidth}
        \centering
        \includegraphics[width=\linewidth]{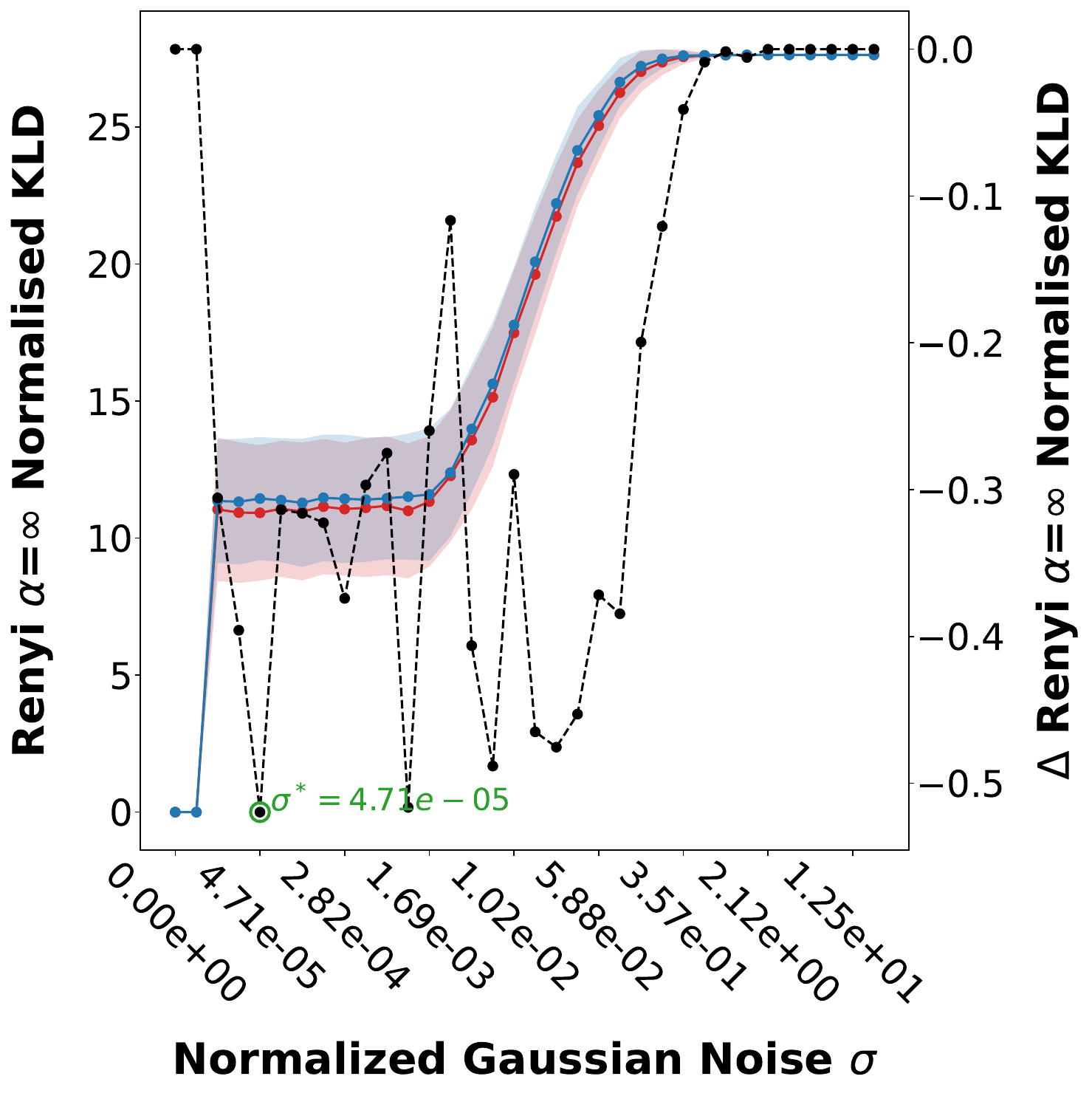}
        \caption{Renyi $\alpha=\infty$ Normalised KL-Divergence}
    \end{subfigure}\hfill
    \begin{subfigure}{0.32\textwidth}
        \centering
        \includegraphics[width=\linewidth]{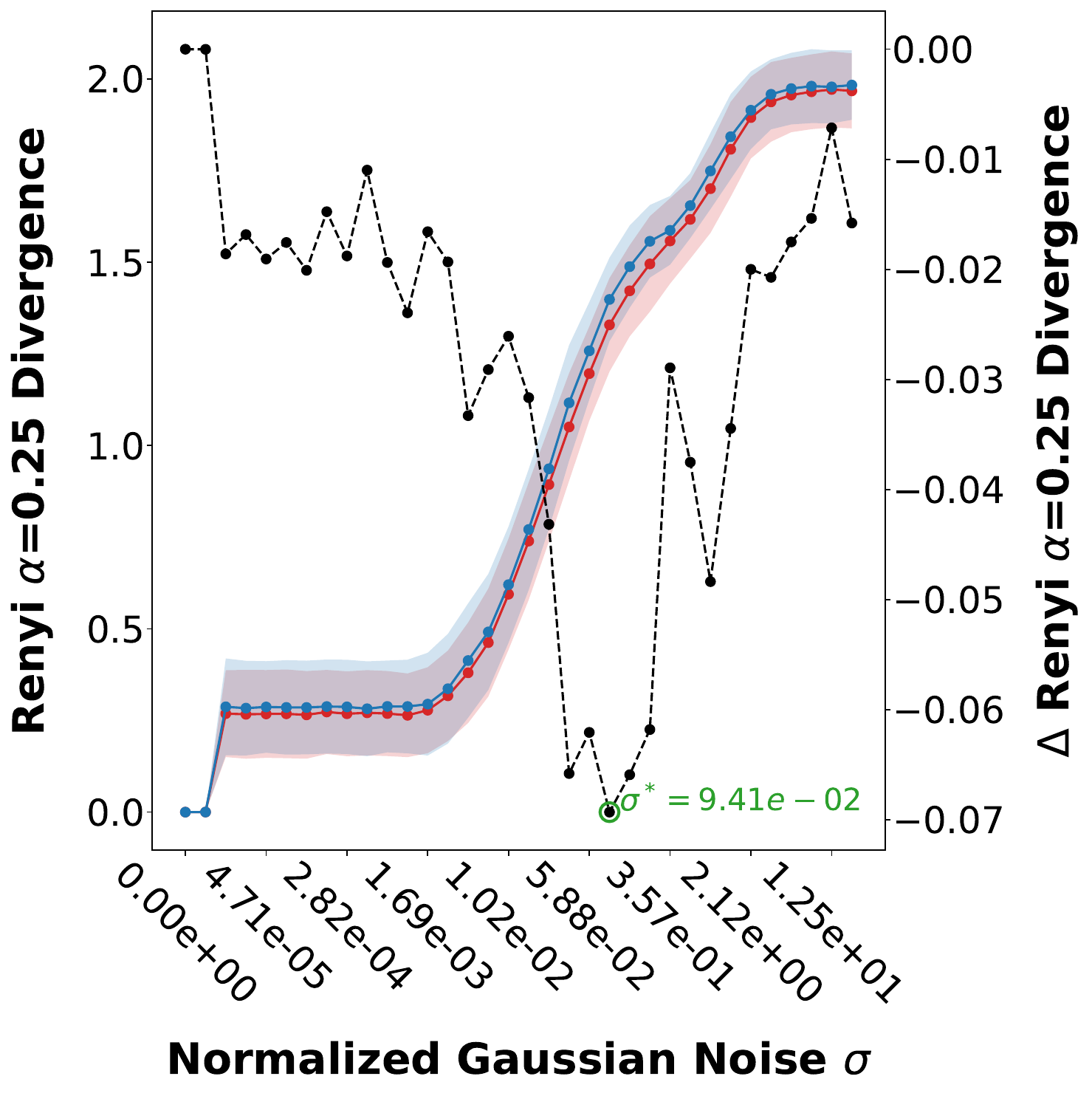}
        \caption{Renyi $\alpha=0.25$ Divergence}
    \end{subfigure}

    \vspace{0.5em}

    \begin{subfigure}{0.32\textwidth}
        \centering
        \includegraphics[width=\linewidth]{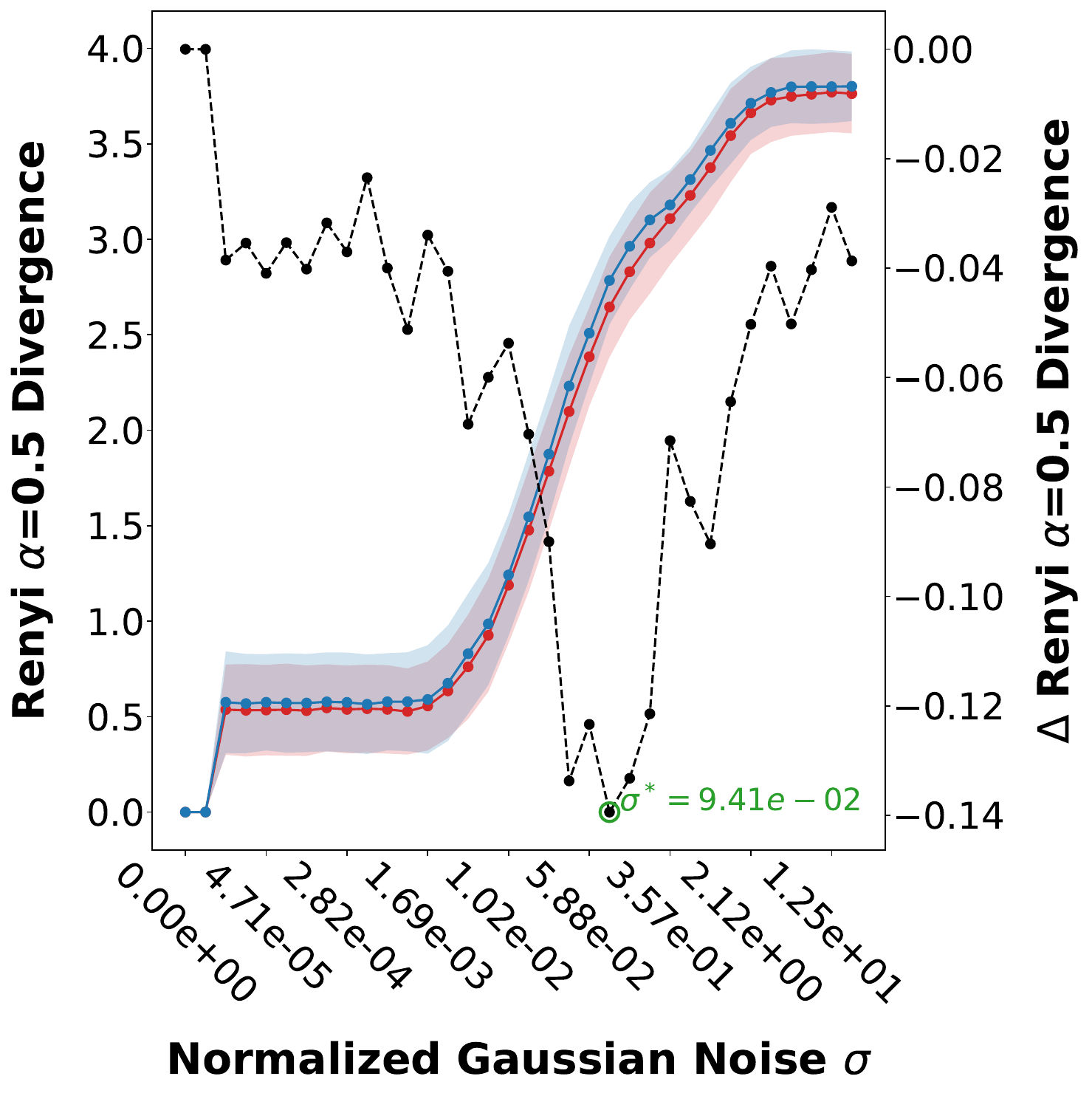}
        \caption{Renyi $\alpha=0.5$ Divergence}
    \end{subfigure}\hfill
    \begin{subfigure}{0.32\textwidth}
        \centering
        \includegraphics[width=\linewidth]{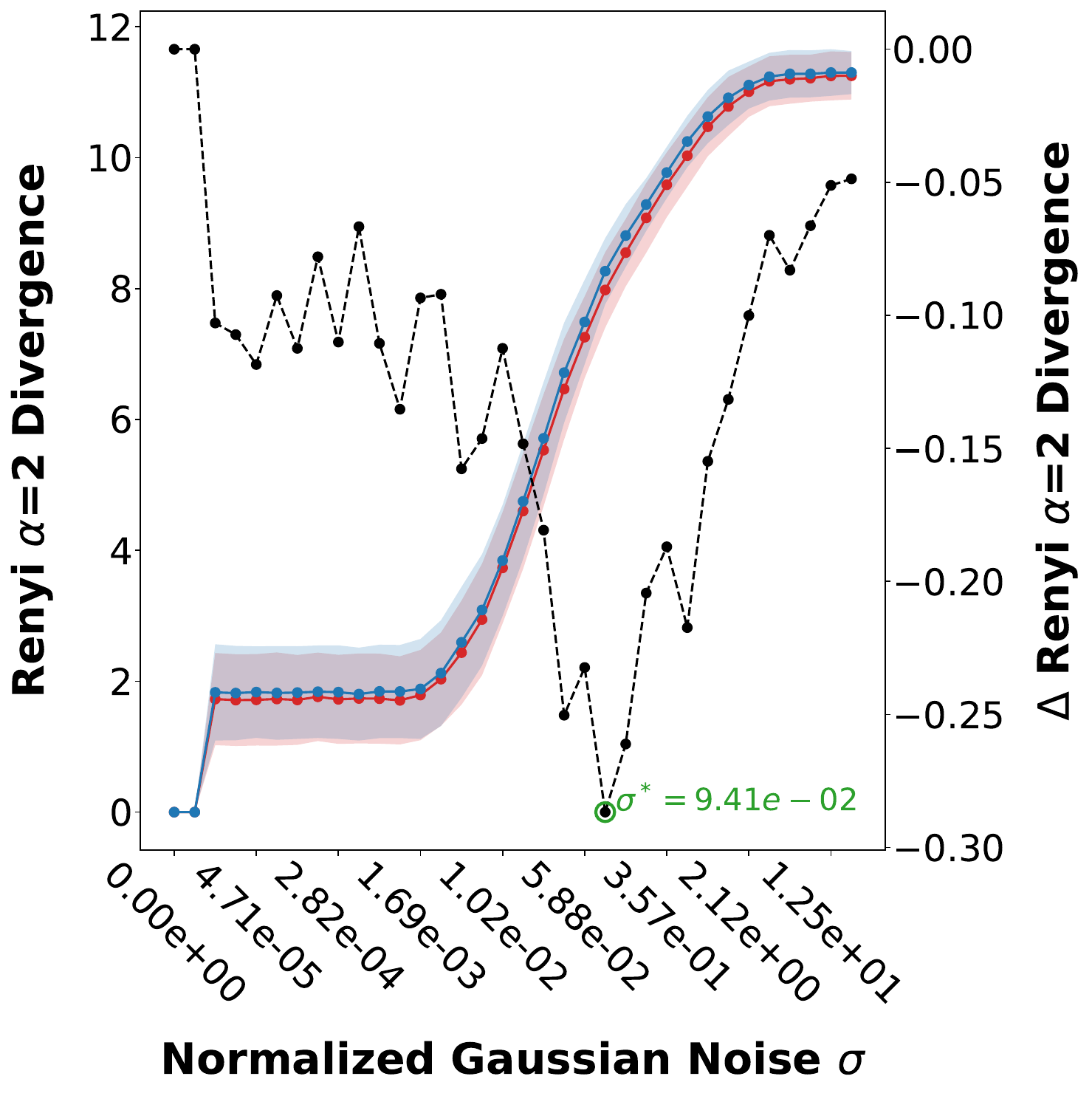}
        \caption{Renyi $\alpha=2$ Divergence}
    \end{subfigure}\hfill
    \begin{subfigure}{0.32\textwidth}
        \centering
        \includegraphics[width=\linewidth]{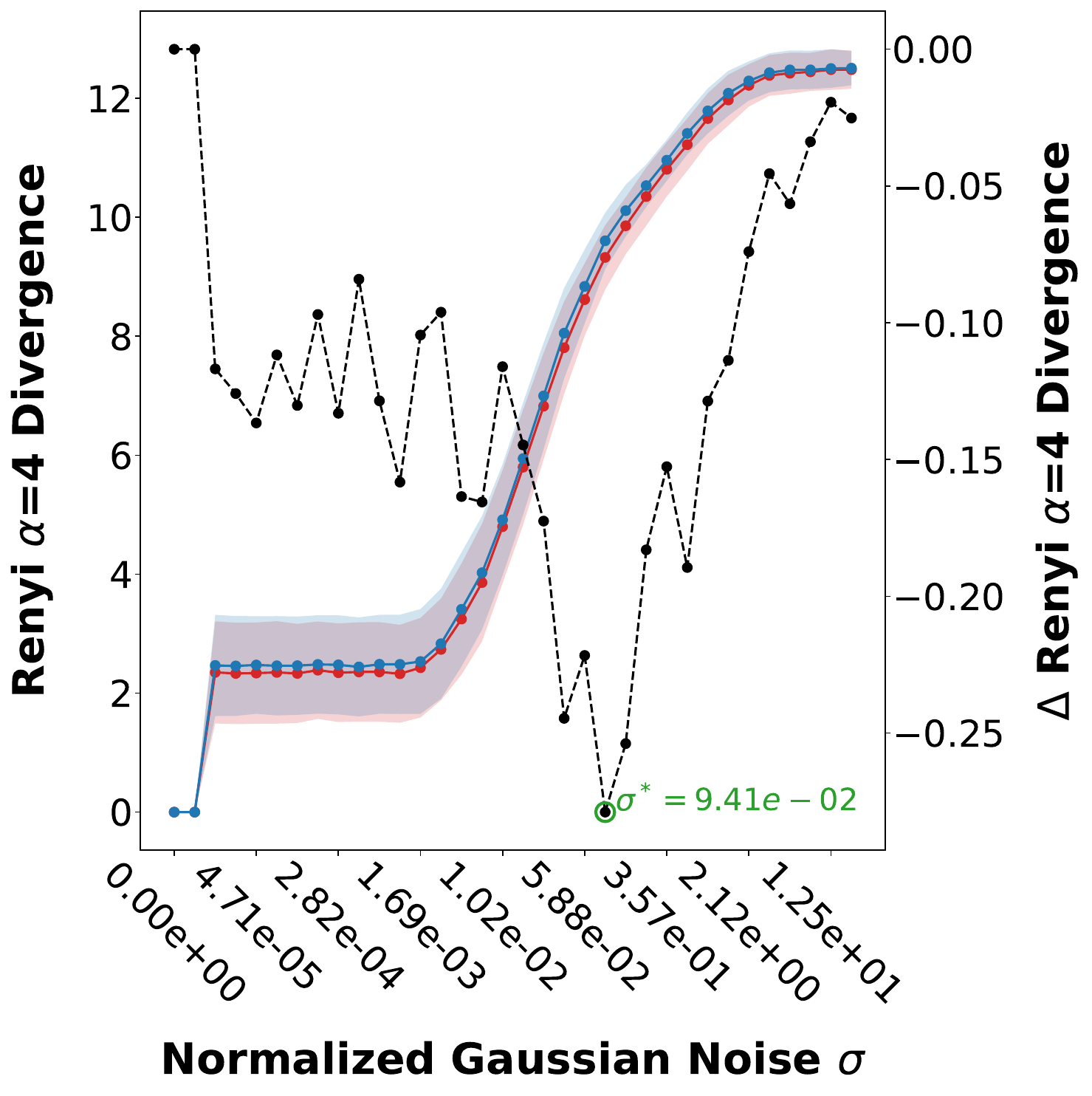}
        \caption{Renyi $\alpha=4$ Divergence}
    \end{subfigure}

    \caption{Raw divergence curves for all tested normalization for a endoscopy target subset members (red), non-members (blue), and member-non-member difference (black) against the Hulu-Med-32B model.}
    \label{fig:raw_divs_hulu_endo}
\end{figure*}

Expanding on \autoref{fig:kld_dynamics}, we additionally provide the divergence dynamics plots for all target models (Figure \ref{fig:raw_divs_llava}, Figure \ref{fig:raw_divs_minigpt}, and Figure \ref{fig:raw_divs_hulu_endo}). The standard deviation within each class is shown as a shaded band around the corresponding curve.

